\documentclass[10pt,journal,compsoc]{IEEEtran}
\usepackage{amsmath,amsfonts}
\usepackage{algorithmic}
\usepackage{algorithm}
\usepackage{array}
\usepackage[caption=false,font=normalsize,labelfont=sf,textfont=sf]{subfig}
\usepackage{textcomp}
\usepackage{stfloats}
\usepackage{url}
\usepackage{verbatim}
\usepackage{bbding}
\usepackage{graphicx}
\usepackage{cite}
\usepackage{booktabs} %
\usepackage{multirow}
\usepackage{tabularx}
\usepackage{wrapfig}
\usepackage{mathtools}
\usepackage{lipsum}
\usepackage{footmisc}
\usepackage[dvipsnames]{xcolor}
\usepackage{xspace}
\usepackage{ragged2e}
\begin{document}

\def \cf {{cf.\thinspace}}
\def \etal {{\emph{et al}.\thinspace}}
\def \eg {{\emph{e.g.}\thinspace}}
\def \ie {{\emph{i.e.}\thinspace}}
\def \vs {{\emph{v.s.}\thinspace}}
\def \etc {{etc.\ }}

\newcommand{\jy}[1]{#1}
\newcommand{\TODO}[2]{{\color{red}#1}}

\newcommand{\name}{hierarchical neural semantic representation\xspace}
\newcommand{\abb}{HNSR\xspace}

\newcommand{\abbnew}{CHNSR\xspace}
\newcommand{\namenew}{clean hierarchical neural semantic representation\xspace}
\newcommand{\newmethod}{self-supervised feature refinement mechanism\xspace}


\title{CamFlow+: Hybrid Motion Bases for 2D Camera Motion Estimation with Stabilization Applications}

\author{Haipeng Li, Zhen Liu, Zhanglei Yang, Hai Jiang, Tianhao Zhou,  Zhengzhe Liu, \\ Ping Tan,~\IEEEmembership{Senior Member,~IEEE},  Bing Zeng,~\IEEEmembership{Fellow,~IEEE} and Shuaicheng Liu,~\IEEEmembership{Senior Member,~IEEE}

\IEEEcompsocitemizethanks{
\IEEEcompsocthanksitem Haipeng Li, Zhen Liu, Zhanglei Yang, Bing Zeng and Shuaicheng Liu are with the School of Information and Communication Engineering, University of Electronic Science and Technology of China, Chengdu, Sichuan, China.

\IEEEcompsocthanksitem Hai Jiang is with the School of Aeronautics and Astronautics, Sichuan University, Chengdu, Sichuan, China.

\IEEEcompsocthanksitem Tianhao Zhou is with the YingCai Honors College, University of Electronic Science and Technology of China, Chengdu, Sichuan, China.

\IEEEcompsocthanksitem Zhengzhe Liu is with Lingnan University, Hong Kong, China.

\IEEEcompsocthanksitem Ping Tan is with the Hong Kong University of Science and Technology and Shenzhen Loop Area Institute.

\IEEEcompsocthanksitem Corresponding author: Shuaicheng Liu (liushuaicheng@uestc.edu.cn).\vspace{2em}}

\thanks{This work was supported in part by the National Natural Science Foundation of China (NSFC) under grants No.62506149, No.62372091, the Hainan Province Science and Technology Plan Project under Grant ZDYF2024(LALH)001, the 111 Projects under Grant B17008, and Lingnan University StartUp Grant fund code: 103684, SDS Interdisciplinary and Strategic Research Grant (ISRG) 784008, and Faculty Research Grant fund code:106106 and 106119. }}

\markboth{Journal of \LaTeX\ Class Files,~Vol.~14, No.~8, August~2021}%
{Shell \MakeLowercase{\textit{et al.}}: A Sample Article Using IEEEtran.cls for IEEE Journals}


\IEEEtitleabstractindextext{%
\justifying
\begin{abstract}
Estimating 2D camera motion is fundamental to computer vision and computational photography.
Existing homography-based methods work well for planar scenes or pure rotation, but struggle with camera translation, depth variation, and local parallax; local homography and mesh-based models improve flexibility but still rely on piecewise planar assumptions.
We introduce CamFlow+, a hybrid-basis framework that represents 2D camera motion directly in dense-flow space.
CamFlow+ combines homography-derived physical bases, stochastic bases sampled from homography flows, and depth-translational bases derived from depth and camera intrinsics, relaxing the single-plane constraint while preserving camera-motion regularity.
A depth-aware smoothness term further regularizes translation-induced parallax in continuous-depth regions while preserving motion changes near depth boundaries.
We evaluate CamFlow+ on GHOF-Cam, a camera-motion benchmark that masks out dynamic objects and ill-posed occlusion regions in an optical-flow benchmark to isolate camera-induced motion.
Experiments show that CamFlow+ improves sparse and dense camera-motion estimation.
In digital video stabilization, CamFlow+ also improves global and local stability, achieving the best top-$1$ preference rate in a blind user study.
Code and datasets will be available on the project page: \url{https://lhaippp.github.io/CamFlow+}.
\end{abstract}

}

\maketitle

\begin{figure*}[t]
    \centering
    \includegraphics[width=1\linewidth]{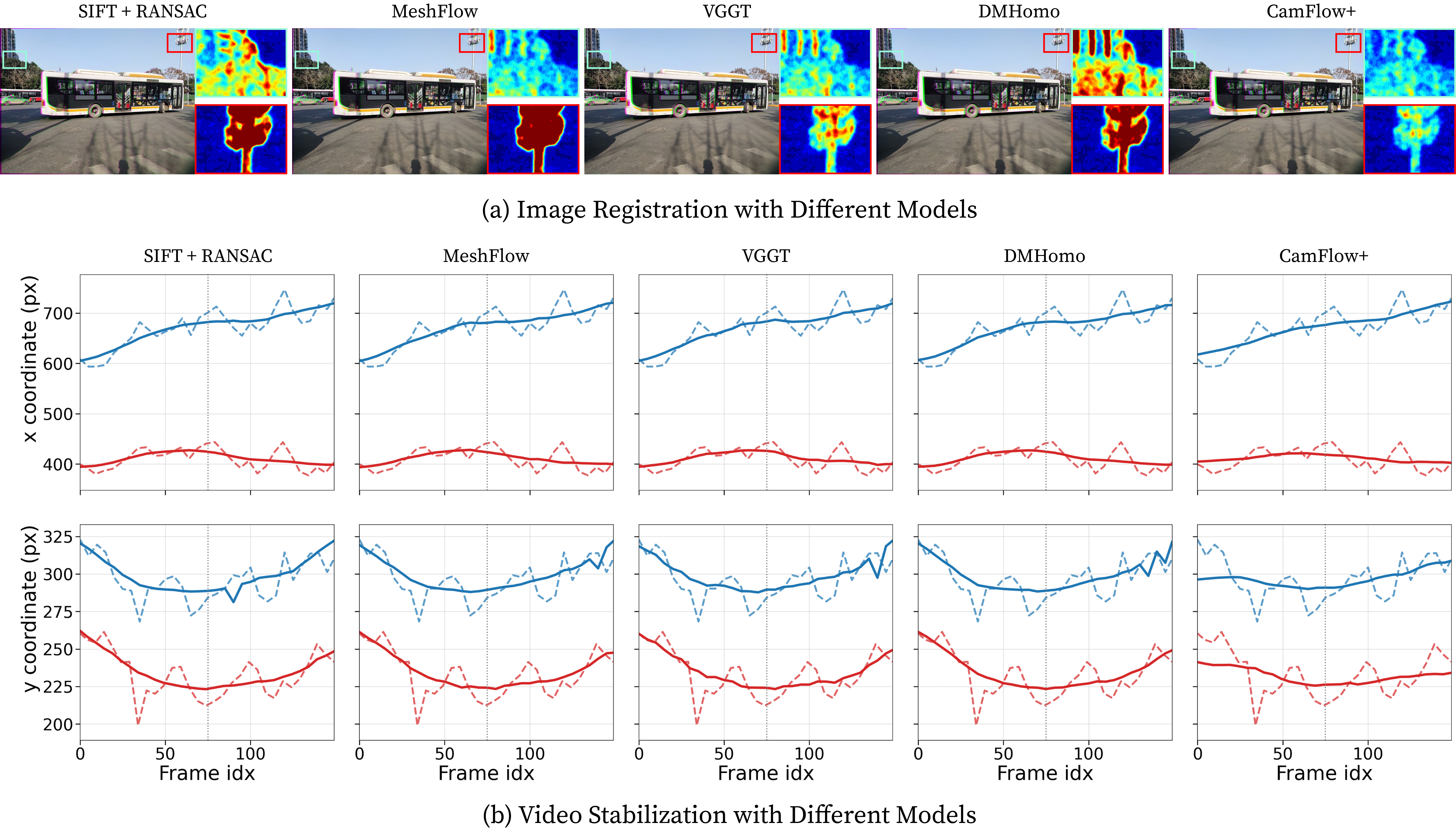}
    \caption{Results of camera-motion estimation for image registration and video stabilization. (a) We compare CamFlow+ with representative single-plane, multi-plane, geometric foundation, and recent homography baselines. Red-channel overlays and error heat maps visualize registration quality; weaker color ghosting and fewer red regions indicate better alignment. (b) The plots show the $x$- and $y$-coordinates of tracked background (red) and foreground (blue) points before stabilization (dashed) and after stabilization (solid). CamFlow+ produces more stationary stabilized trajectories with reduced residual motion and high-frequency jitter, leading to better video stabilization.}
    \label{fig:Teaser}
\end{figure*}


\section{Introduction}
\label{intro}
\IEEEPARstart{C}{amera} motion estimation is a basic building block in computer vision and computational photography, and it is important for mobile imaging, where hand-held capture makes frame-to-frame motion unavoidable. Many imaging pipelines rely on accurate motion estimation, including stereo matching~\cite{lucas1981iterative}, panoramic image stitching~\cite{brown2007automatic}, multi-frame image and video enhancement~\cite{xue2019video,hasinoff2016burst}, video deblurring~\cite{su2017deep}, and digital video stabilization~\cite{liu2013bundled,LiuTYSZ16}.

In this paper, we focus on \emph{2D camera motion}, namely the image-plane displacement induced by 3D camera motion after projection through scene geometry and camera intrinsics~\cite{DLT}. This representation is directly consumed by image alignment, warping, and stabilization, which operate on pixels. A homography is a compact special case of 2D camera motion: when the scene is planar, or when the camera undergoes pure rotation, the projection from one view to another can be written as
\begin{equation}
\mathbf{H} = \mathbf{K} \left( \mathbf{R} + \mathbf{t} \frac{\mathbf{n}^T}{d} \right) \mathbf{K}^{-1},
\label{eq:camera_pose_to_homo}
\end{equation}
where \(\mathbf{R}\) and \(\mathbf{t}\) denote camera rotation and translation, \(\mathbf{n}\) and \(d\) denote the normal and distance of the scene plane, and \(\mathbf{K}\) is the intrinsic matrix. This formulation also indicates how to move beyond a single homography. In real scenes with multiple planes or non-planar structures, the effect of camera translation varies with scene geometry and depth. The camera-induced motion is therefore better described as a dense, depth-dependent, non-linear flow field.

{Existing camera-motion representations can be broadly organized into two families. The first family uses compact global models. Classical feature-based pipelines estimate the homography from matched keypoints~\cite{sift}, while learning-based methods regress homography parameters from image pairs~\cite{CA-Unsupervised2020,HomoGAN2022}. BasesHomo~\cite{BasesHomo2021} further reformulates homography estimation as a linear combination of motion bases, providing a useful basis-space view of camera motion. These methods are reliable in dominant-plane or near-pure-rotation cases, but have difficulty handling multi-plane scenes and depth-dependent local variation.}

{The second family uses multiple-plane or local motion models. MeshFlow~\cite{LiuTYSZ16} partitions the image into a grid, estimates local motions, and smooths them into a dense motion field; its learning-based variants further estimate multi-resolution meshes or mesh bases from image pairs~\cite{liu2022content,liu2023unsupervised}. Related local-homography and homography-mixture methods similarly increase expressiveness by fitting multiple local planar transforms~\cite{grundmann2012calibration,shen2020ransac}. Such models are more flexible than a single homography and have been widely used in video stabilization~\cite{liu2013bundled,LiuTYSZ16}. However, in challenging videos, dynamic objects, depth discontinuities, and near-far motion differences can contaminate local estimates, while spatial smoothing may introduce broken structures, line bending, or geometric distortion.}


These limitations motivate a representation that is more flexible than a single homography or a set of local planar models, while still retaining camera-motion regularity. The conference version, CamFlow, addresses this goal by representing 2D camera motion as a weighted combination of dense motion bases, so the final motion field can capture both global structure and local variation. Based on this idea, CamFlow combines homography-derived physical bases, obtained from a second-order Taylor expansion of projective motion, with stochastic bases extracted from randomly sampled homography flows to cover more complex residual patterns. It also uses a probabilistic motion loss to stabilize training. It also introduced GHOF-Cam, a benchmark derived from optical flow benchmark GHOF~\cite{gyroflow+} by masking dynamic objects and ill-posed occlusion regions so that the remaining flow isolates camera-induced motion.

This journal version extends CamFlow into \textbf{CamFlow+} in both representation design and application scope. First, we introduce {depth-translational bases}, which use depth and camera intrinsics to derive analytic translation-induced flows along the camera axes. These bases transform the original depth-agnostic hybrid basis space into a depth-aware representation that explicitly models the depth-dependent parallax pattern in which nearer regions undergo larger apparent motion than farther regions. We further regularize the estimated motion field with depth-aware smoothness, encouraging coherent motion in continuous-depth regions while allowing changes at depth boundaries. By explicitly modeling parallax with depth-translational bases, CamFlow+ provides a more structured motion representation and achieves the best average performance under GHOF/GHOF-Cam camera-motion evaluation protocols. Fig.~\ref{fig:Teaser}(a) gives a visual example. In the red-channel overlay, weaker color ghosting indicates better alignment; in the heat map, red regions indicate larger errors. Compared with SIFT+RANSAC~\cite{sift,ransac} and MeshFlow~\cite{LiuTYSZ16} as representative traditional single- and multi-plane baselines, VGGT~\cite{wang2025vggt} as a geometric foundation model, and DMHomo~\cite{li2024dmhomo} as a recent SOTA homography method, CamFlow+ produces cleaner alignment and fewer local errors.

Second, we extend CamFlow+ to digital video stabilization (DVS), a practical application where camera-motion errors directly appear as residual jitter, wobble, or depth-boundary distortion. In the stabilization pipeline, we replace only the camera-motion estimation module while keeping other stages unchanged, making the final stabilization quality a direct test of the estimated camera motion. Fig.~\ref{fig:Teaser}(b) visualizes stabilization quality through point trajectories: a desirable result should make the camera path smooth while keeping both background and foreground points stable. CamFlow+ achieves better global and local stabilization, reducing both low-frequency drift and high-frequency shake. This application-level benefit is also supported by more detailed comparisons and a blind user study: CamFlow+ receives 43.30\% top-$1$ preference, ahead of the second-ranked HM-Mix~\cite{grundmann2012calibration} at 26.79\%.
Our main contributions are:
\begin{itemize}
    \item We propose a hybrid basis representation that models complex 2D camera motion in the dense flow space, relaxing the single-plane constraint of homography-based formulations.
    \item We introduce depth-translational bases and depth-aware smoothness to explicitly model translation-induced parallax with depth guidance.
    \item We conduct comprehensive experiments, including sparse, dense, perceptual, and application-level studies, on GHOF/GHOF-Cam to validate CamFlow+ under both benchmark and stabilization settings.
    \item We extend camera-motion evaluation to video stabilization, demonstrating that improved camera-motion estimation yields practical stabilization gains.
\end{itemize}


In comparison to our earlier conference version, we make the following substantial new contributions:
\begin{itemize}
    \item We provide a deeper analysis of the basis-addition mechanism, showing that direct flow addition relaxes exact homography composition and thereby increases expressivity in the dense flow space.
    \item We extend the original CamFlow representation to CamFlow+ by introducing depth-translational bases derived from depth and camera intrinsics, enabling explicit modeling of translation-induced parallax.
    \item We further introduce a depth-aware smoothness regularizer for the estimated motion field, encouraging coherent motion within continuous-depth regions while preserving changes at depth boundaries.
    \item We introduce digital video stabilization as an application-level evaluation of 2D camera-motion estimation, including diverse qualitative comparisons and a blind user study that verifies the practical benefit of the proposed depth-aware representation.
    \item We extend the evaluation on the previously introduced GHOF-Cam benchmark with depth-aware CamFlow+, 3D foundation models, optical-flow baselines, and application-level DVS results.
\end{itemize}


\section{Related Work}
\label{subsec:relate_work}

{
\subsection{Single Plane Methods}
Traditional single-plane methods mainly rely on homography, the estimation typically follows three stages: feature detection using algorithms such as SIFT~\cite{sift} or ORB~\cite{orb}, correspondence matching~\cite{cunningham2021k}, and outlier rejection techniques like RANSAC~\cite{ransac}. Recent advances in learning-based feature detection and matching, including LIFT~\cite{lift}, SuperPoint~\cite{superpoint}, and SOSNet~\cite{sosnet}, have improved the homography estimation. Additionally, enhanced outlier rejection methods such as MAGSAC~\cite{magsac} and MAGSAC++~\cite{barath2020magsac++} have increased stability in challenging scenarios involving multiple planes, parallax, and dynamic foregrounds.

Optimization-based approaches~\cite{Clkn,DPCP}, such as those derived from Lucas-Kanade or sum of squared differences, iteratively refine homography parameters from an initial estimate. Deep learning methods have further advanced homography estimation, beginning with supervised approaches~\cite{supervised2016} that rely on synthetic image pairs. More recent methods can be categorized into supervised~\cite{Dynamic-supervised2020,Crossresolution-supervised2021,Iterative-supervised2022,jiang2023supervised,LBHomo,jiang2026supervised,liu2024codinghomo} and unsupervised~\cite{unsupervised2018,unsupervised2020,BasesHomo2021,HomoGAN2022} frameworks. Along the supervised direction, realistic and diffusion-based data generation has been used to improve both small-baseline and large-baseline homography learning~\cite{jiang2023supervised,LBHomo,jiang2026supervised}, while video-coding priors have also been exploited to bootstrap deep homography estimation~\cite{liu2024codinghomo}. Unsupervised techniques have gained popularity due to their label-free training strategies. For example, CAHomo~\cite{CA-Unsupervised2020} utilizes a self-guided mask to highlight key feature points, while BasesHomo~\cite{BasesHomo2021} constrains the rank of feature maps by learning an 8-dimensional motion basis for improved estimation. HomoGAN~\cite{HomoGAN2022} introduces a Generative Adversarial Network (GAN) loss to identify the dominant plane and integrates a Transformer encoder for coarse-to-fine refinement, DMHomo~\cite{li2024dmhomo} leverages diffusion model to generate supervised image pairs to bootstrap the performance. Additionally, SCPNet~\cite{zhang2024scpnet} and McNet~\cite{zhu2024mcnet} explore cross-modal homography estimation.

Despite these advancements, homography remains a single-plane parametric model, limiting its ability to fully represent complex camera motion. To overcome this, we introduce a hybrid motion-basis representation designed to model multi-plane, non-linear motion effectively.

\subsection{Multi-plane Methods}
Mesh-based warping is widely used for multi-plane scenes, where each grid cell or image region is assigned a local motion model. Representative methods include dual-homography models~\cite{gao2011constructing}, APAP~\cite{zaragoza2013projective}, Bundled Paths~\cite{liu2013bundled,igarashi2005rigid}, HM-Mix~\cite{grundmann2012calibration}, and MeshFlow~\cite{liu2016meshflow}, which increase flexibility by combining local projective warps with spatial smoothing.
Learning-based mesh methods~\cite{liu2022content,liu2023unsupervised,liu2024unsupervised} further estimate multi-resolution meshes or grid-wise motion bases. However, these representations still depend on local homography assumptions and grid-wise optimization, limiting their ability to represent complex depth-dependent camera motion.

Optical flow provides a dense per-pixel alternative and is naturally expressive for multi-plane motion. Methods such as FlowNet~\cite{dosovitskiy2015flownet}, SpyNet~\cite{ranjan2017spynet}, PWC-Net~\cite{sun2018pwcnet}, RAFT~\cite{teed2020raft}, FlowFormer~\cite{huang2022flowformer}, and Sea-RAFT~\cite{wang2024sea} have achieved strong correspondence estimation. Nevertheless, generic optical flow also captures independently moving objects and occlusion artifacts, which conflicts with pure camera-motion estimation.
Generative priors have also been explored for geometric motion and warping tasks, including diffusion-based image rectangling for stitching~\cite{zhou2024recdiffusion} and repurposed diffusion image priors for motion estimation~\cite{wang2025stablemotion}. Sensor-assisted methods use inertial measurements as complementary motion cues, such as gyroscope-guided optical-flow and homography learning~\cite{GyroFlow,gyroflow+} and gyroscope-guided optical image stabilizer compensation~\cite{9509028}.
3D geometric foundation models provide another route by predicting 3D assets from which pose, intrinsics, and depth can derive 2D camera motion. DUSt3R~\cite{dust3r_cvpr24}, VGGT~\cite{wang2025vggt}, and $\pi^3$~\cite{wang2025pi3} are representative examples, while related 3D correspondence and registration studies further use foundation-model semantic representations~\cite{du2025hierarchical} or probabilistic masks for moving-object-aware point-cloud registration~\cite{du2025hybridreg}. However, under our GHOF-Cam protocol, feed-forward 3D geometry models still show limited zero-shot performance for 2D camera-motion estimation.

CamFlow+ uses depth and camera intrinsics to construct depth-translational 2D motion bases, and linearly combines them with homography-derived and stochastic bases to model multi-plane camera motion in dense-flow space. This hybrid basis formulation avoids grid-wise mesh optimization, reduces sensitivity to dynamic-object and occlusion artifacts in generic optical flow, and achieves stronger 2D camera-motion accuracy than feed-forward 3D geometric foundation models under the GHOF-Cam protocol.
}

\section{Method}

\begin{figure}[t]
    \centering
    \includegraphics[width=1\linewidth]{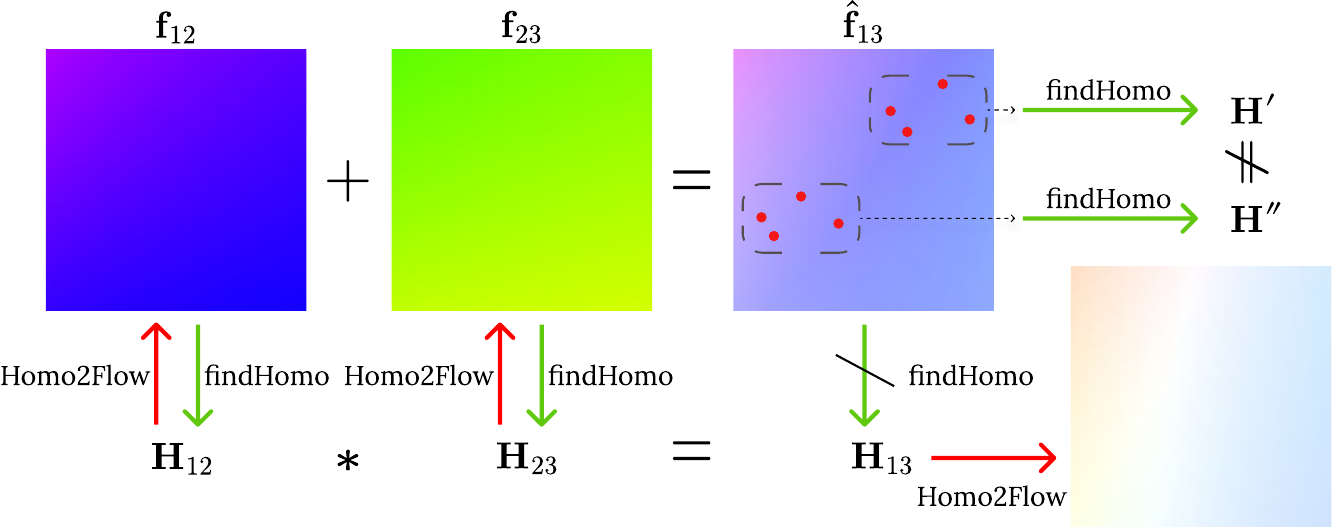}
    \caption{{Intuition of basis addition. Two homographies induce two flow fields. Exact composition keeps the result within the homography family, while direct flow addition produces a composite dense flow. Homographies fitted from different point samples of this additive flow become inconsistent, showing that flow addition relaxes strict homography-induced warping.}}
    \label{fig:Motion-Basis}
\end{figure}

\subsection{Motion Basis}
\label{subsec:motion_basis}
{Unlike geometrically strict homography formulations such as CAHomo~\cite{CA-Unsupervised2020}, which predict four corner offsets (i.e., the 4-point representation~\cite{DLT}) and then solve for a homography, BasesHomo~\cite{BasesHomo2021} represents camera motion as an additive combination of multiple motion bases. The results in Table~\ref{tab:Compare_CAunsup} and Table~\ref{tab:GHOF-Cam-Quan} indicate that this representation is beneficial for both globally dominant camera motion and locally varying multi-plane motion. In this section, we analyze the mechanism behind basis addition and, based on this analysis, develop a hybrid basis space composed of physical and stochastic bases.}

\subsubsection{{Intuition of Basis Addition.}} {We explain this intuition using homography as an example. Let \(\phi(\mathbf{H})\) denote the operation that converts a homography into its induced dense flow, obtained by warping a regular grid map with \(\mathbf{H}\) and computing the displacement as the warped grid minus the original grid. Consider three consecutive views, where \(\mathbf{H}_{12}\) denotes the homography-induced motion from view 1 to view 2, and \(\mathbf{H}_{23}\) denotes the motion from view 2 to view 3. We first obtain two flows \(\mathbf{f}_{12}=\phi(\mathbf{H}_{12})\) and \(\mathbf{f}_{23}=\phi(\mathbf{H}_{23})\). If we want to compose these two motions into the motion from view 1 to view 3 in a geometrically accurate manner, we should compose them in the flow space with remapping:}
\begin{equation}
{\begin{aligned}
\mathbf{x}' &= \mathbf{x} + \mathbf{f}_{12}(\mathbf{x}), \qquad
\mathbf{f}_{13}(\mathbf{x}) = \mathbf{f}_{12}(\mathbf{x}) + \mathbf{f}_{23}(\mathbf{x}').
\end{aligned}}
\end{equation}
{where \(\mathbf{f}_{13}\) corresponds to the accurately composed homography-induced flow. This composition is accurate because the second motion is evaluated at the remapped starting point rather than at the original pixel location.
}
{The key difference of basis addition is that it no longer enforces the strict warping relationship defined by exact homography composition. Instead of using the remapped composition above, it adopts the simpler form}
\begin{equation}
{\mathbf{f}_{13}(\mathbf{x}) \approx \hat{\mathbf{f}}_{13}(\mathbf{x}) = \mathbf{f}_{12}(\mathbf{x}) + \mathbf{f}_{23}(\mathbf{x}).}
\end{equation}
{This form directly adds the two flows at the same pixel location, instead of querying \(\mathbf{f}_{23}\) at the remapped position \(\mathbf{x}'\). Its relation to exact flow composition can be seen from a local Taylor expansion. When \(\mathbf{f}_{23}\) is locally smooth and the displacement \(\mathbf{f}_{12}(\mathbf{x})\) is small,}
\begingroup
\small
\begin{equation}
{\mathbf{f}_{23}(\mathbf{x}+\mathbf{f}_{12}(\mathbf{x}))
= \mathbf{f}_{23}(\mathbf{x})
+ \nabla \mathbf{f}_{23}(\mathbf{x})\mathbf{f}_{12}(\mathbf{x})
+ \mathcal{O}(\|\mathbf{f}_{12}\|^2).}
\end{equation}
\endgroup
{Here \(\nabla \mathbf{f}_{23}(\mathbf{x})\) denotes the spatial Jacobian of \(\mathbf{f}_{23}\) at \(\mathbf{x}\), and \(\mathcal{O}(\|\mathbf{f}_{12}\|^2)\) denotes the remaining terms whose magnitude is second order or higher in the displacement \(\mathbf{f}_{12}\). The additive form keeps \(\mathbf{f}_{23}(\mathbf{x})\) and drops the Jacobian interaction term and higher-order terms. Here \(\mathbf{f}_{23}\) is naturally defined on the second-view coordinate domain; evaluating it at \(\mathbf{x}\) instead of \(\mathbf{x}'\) intentionally relaxes this domain consistency, allowing the additive flow \(\hat{\mathbf{f}}_{13}\) to deviate from the accurately composed flow \(\mathbf{f}_{13}\).}

{As shown in Fig.~\ref{fig:Motion-Basis}, we first generate two homography-induced flows and add them as \(\hat{\mathbf{f}}_{13}(\mathbf{x})=\mathbf{f}_{12}(\mathbf{x})+\mathbf{f}_{23}(\mathbf{x})\). We then randomly sample point correspondences from \(\hat{\mathbf{f}}_{13}\) and fit homographies from different subsets, i.e., $\mathbf{H}', \mathbf{H}''$. If the additive flow were induced by a single exact homography, any non-degenerate point subset would recover the same homography up to numerical error. The fitted homographies are instead inconsistent, which indicates that the additive flow is not constrained to a single exact homography-induced warp. This observation suggests that direct flow addition can represent more complex motion patterns beyond strict homography composition. This interpretation is also supported by the ablation study in BasesHomo~\cite{BasesHomo2021}, where replacing the 4-point representation with a motion-basis representation under the same network architecture improves the average error.}
{Motivated by this insight, our next step is to design a motion-basis framework for modeling 2D camera motion under the flow-addition representation. To this end, we introduce a hybrid basis formulation that combines physical bases and stochastic bases, so that the model can jointly capture global and local geometry.}

\begin{figure*}[t]
    \centering
    \includegraphics[width=1\linewidth]{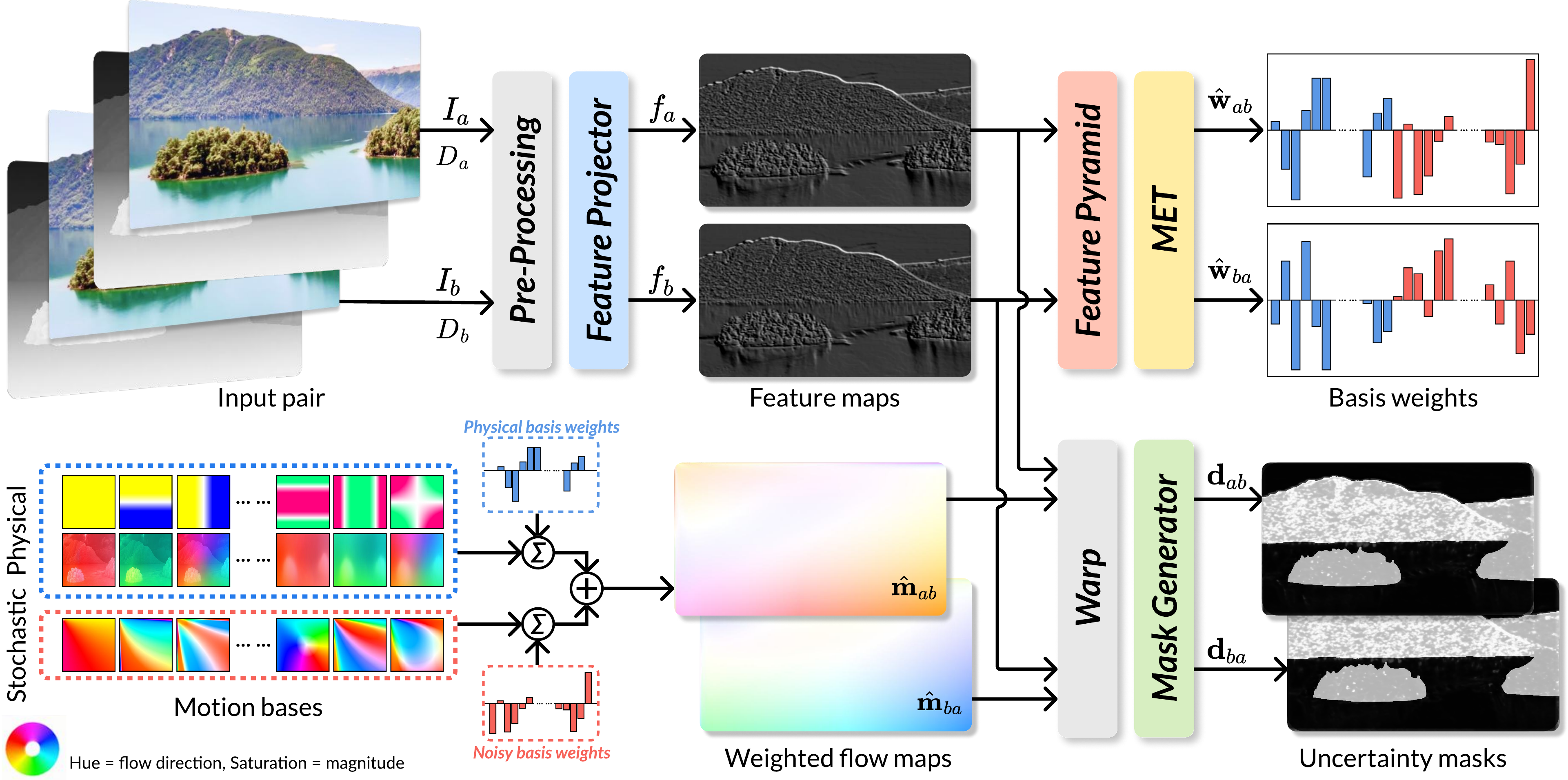}
    \caption{Our proposed motion estimation framework. Given image pair \((I_a, I_b)\), features are extracted through a multi-scale pyramid and processed by the motion estimation transformer (MET) to compute weights for physical (blue) and stochastic (noisy, red) motion bases. These weights linearly combine predefined motion bases to generate flow maps for warping. A mask generator predicts confidence maps \(\mathbf{d}_{ab}\) and \(\mathbf{d}_{ba}\) to reject unreliable regions, enhancing estimation robustness.}
    \label{fig:network-ppl}
\end{figure*}

\subsubsection{{Physical Bases Family.}} {Despite the improved performance of motion-basis representations, the bases in BasesHomo~\cite{BasesHomo2021} are mainly derived from homography-induced motion and remain depth-agnostic, which limits their ability to model stronger parallax. We therefore define the physical basis family using two complementary components: homography bases for depth-agnostic low-order geometry, and depth-translational bases for depth-aware parallax induced by camera translation.}

\noindent\textbf{{Homography Bases.}} {We first retain a homography bases branch so that the representation captures the dominant low-order structure of homography-induced motion.} Consider a pixel with homogeneous coordinates $\mathbf{P}(x, y) = [x, y, 1]^T \!\in\! \mathbb{R}^3$. The homography maps it to a homogeneous vector, which is projected back to image coordinates:
\begin{equation}
\tilde{\mathbf{P}}' = \mathbf{H}\mathbf{P}(x,y), \qquad
\mathbf{p}' =
\left[
\frac{\tilde{P}'_1}{\tilde{P}'_3},
\frac{\tilde{P}'_2}{\tilde{P}'_3}
\right]^T.
\end{equation}
The induced 2D motion is \(\mathbf{p}'-[x,y]^T=[\Delta x,\Delta y]^T\), where
\begin{equation}
\begin{split}
\Delta x &= \frac{h_1 x + h_2 y + h_3}{h_7 x + h_8 y + 1} - x, \\
\Delta y &= \frac{h_4 x + h_5 y + h_6}{h_7 x + h_8 y + 1} - y,
\end{split}
\end{equation}
where \( h_1, \dots, h_8 \) denote the elements of matrix \(\mathbf{H}\), and \( h_9 \) is constrained to 1.
{We use normalized image coordinates \(x,y\in[-1,1]\), with the image center as the origin. Unlike the first-order basis used in BasesHomo~\cite{BasesHomo2021}, we derive homography bases from a second-order Taylor expansion, which provides a better local approximation of homography-induced motion under this coordinate system.} For instance, expanding $\Delta x$ around the origin $(0, 0)$ gives:
\begin{align}
\Delta x &= \frac{(h_1 - 1) x + h_2 y + h_3 - h_7 x^2 - h_8 x y}{h_7 x + h_8 y + 1}
\\
&\approx w_1 \cdot 1 + w_2 \cdot x +w_{3} \cdot y + w_4 \cdot xy \\
& \quad  + w_5 \cdot x^{2} + w_6  \cdot y^{2} + \Delta ,
\end{align}
where $w_i,i\in[1,6]$ are coefficients, the basis functions are $b=[1, x, y, xy, x^2, y^2]$ and $\Delta$ denotes the higher-order infinitesimal.
Similarly, $\Delta y$ can be decomposed into this subspace. By combining the decompositions of both $\Delta x, \Delta y$, the homography basis set is represented as
\begin{equation}
\mathcal{B}_{\mathrm{homo}}
=\left\{\left(b_i, 0\right) \mid b_i \in b\right\}
\cup
\left\{\left(0, b_i\right) \mid b_i \in b\right\},
\end{equation}
where $b_i$ denotes the $i$-th element in $b$. {We use these 12 bases in the homography bases branch to represent depth-agnostic low-order motion, which covers dominant-plane motion and minor local deformation.} Each basis can be transformed into an optical flow according to the image coordinate, as shown in Fig.~\ref{fig:network-ppl} (1st row of Physical motion bases). However, they do not explicitly encode depth-dependent parallax, which motivates the depth-translational bases.

\begin{figure}[t]
    \centering
    \includegraphics[width=1\linewidth]{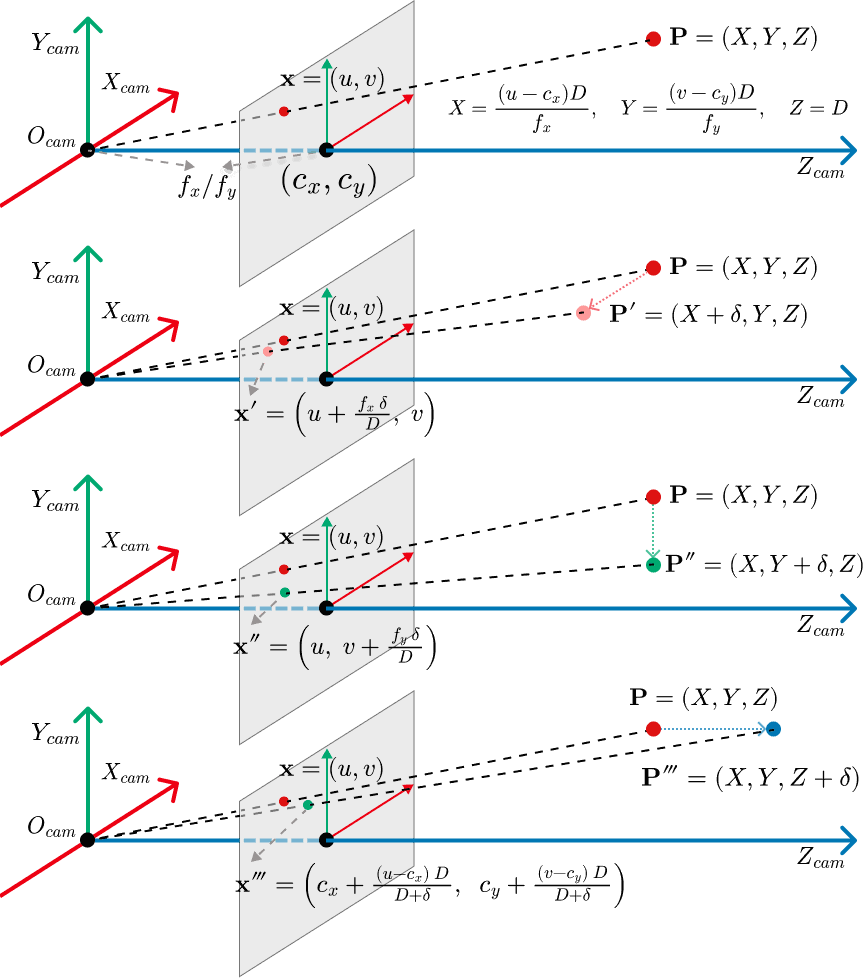}
    \caption{{Illustration of depth-translational bases. Given a depth map, each pixel $\mathbf{x}$ is first unprojected to a 3D camera-coordinate point $\mathbf{P}=(X,Y,Z)$. Perturbing this point along the three camera axes gives $\mathbf{P}'$, $\mathbf{P}''$, and $\mathbf{P}'''$, whose re-projected displacements form the three analytic flow bases $\{\mathbf{f}^{x},\mathbf{f}^{y},\mathbf{f}^{z}\}$ for lateral ($X_{\text{cam}}$), vertical ($Y_{\text{cam}}$), and forward ($Z_{\text{cam}}$) camera translations.}}
    \label{fig:depth-trans-basis}
\end{figure}

\noindent\textbf{{Depth-Translational Basis.}}
{To further extend the physical basis family, we explicitly model the parallax induced by camera translation under depth variation. This component is motivated by the plane-induced homography relation:}
\begin{equation}
{\mathbf{H}=\mathbf{K}\left(\mathbf{R}+\frac{\mathbf{t}\mathbf{n}^{T}}{d}\right)\mathbf{K}^{-1},}
\end{equation}
{where \(\mathbf{n}\) and \(d\) denote the normal and distance of a scene plane. For a single plane, \(\mathbf{n}\) and \(d\) are constant and the induced motion is a global homography. For a non-planar scene, however, the effective depth and local surface geometry vary spatially, making the camera-induced motion a dense, spatially varying field rather than a single homography. We use this relation only to motivate the coupling between translation and scene depth, and construct our depth-translational bases directly by unprojecting each pixel to a 3D point and reprojecting it after small 3D translations. More specifically, given depth \(D(\mathbf{x})\) and camera intrinsics \(\mathbf{K}=\{f_x,f_y,c_x,c_y\}\), a pixel \(\mathbf{x}=(u,v)\) can be unprojected into camera coordinates as}
{\small
\begin{equation}
X = \frac{(u-c_x)D(\mathbf{x})}{f_x}, \quad
Y = \frac{(v-c_y)D(\mathbf{x})}{f_y}, \quad
Z = D(\mathbf{x}).
\label{eq:unproject}
\end{equation}
}
{which recovers the unique 3D point $\mathbf{P}=(X,Y,Z)$ lying on the viewing ray through $\mathbf{x}$ at depth $D(\mathbf{x})$, as illustrated in the first row of Fig.~\ref{fig:depth-trans-basis}.}

{We implement camera translation in an equivalent fixed-camera formulation by perturbing the unprojected 3D points along each camera axis. Thus, for each unprojected point $\mathbf{P}$, we add a small scalar offset $\delta$ along one camera axis at a time, namely $X_{\text{cam}}$, $Y_{\text{cam}}$, and $Z_{\text{cam}}$:}
\begin{equation}
{\begin{aligned}
\mathbf{P}'   &= (X\!+\!\delta,\,Y,\,Z),\\
\mathbf{P}''  &= (X,\,Y\!+\!\delta,\,Z),\\
\mathbf{P}''' &= (X,\,Y,\,Z\!+\!\delta).
\end{aligned}}
\end{equation}
{These three perturbations represent lateral, vertical, and forward camera translations, respectively (rows 2--4 of Fig.~\ref{fig:depth-trans-basis}). We then project each perturbed point $\mathbf{P}^{\star}=(X',Y',Z')$ back to the image plane by}
\begin{equation}
{u'=f_x\frac{X'}{Z'}+c_x,\qquad v'=f_y\frac{Y'}{Z'}+c_y,}
\label{eq:reproject}
\end{equation}
{and define the induced flow as the 2D displacement $\mathbf{f}(\mathbf{x})=(u'-u,\,v'-v)$. The bases have simple geometric forms.}

\noindent\emph{{(a) Lateral translation ($X_{\text{cam}}$-axis).}} {For $\mathbf{P}'=(X\!+\!\delta,Y,Z)$, only the horizontal 3D coordinate changes. Since the depth remains $Z=D(\mathbf{x})$, re-projecting $\mathbf{P}'$ gives a shifted pixel $\mathbf{x}'$:}
\begin{equation}
{\begin{aligned}
\mathbf{x}' &= \Big(u+\tfrac{f_x\,\delta}{D(\mathbf{x})},\;v\Big),\\
\mathbf{f}^{x}(\mathbf{x}) &= \mathbf{x}'-\mathbf{x}
= \Big(\tfrac{f_x\,\delta}{D(\mathbf{x})},\;0\Big).
\end{aligned}}
\end{equation}
{Thus nearby pixels move more than distant pixels. This directly models the near-large/far-small parallax produced by sideways camera motion.}

\noindent\emph{{(b) Vertical translation ($Y_{\text{cam}}$-axis).}} {For $\mathbf{P}''=(X,Y\!+\!\delta,Z)$, the depth is also unchanged, but the vertical image coordinate shifts. The re-projected pixel $\mathbf{x}''$ and the induced flow are}
\begin{equation}
{\begin{aligned}
\mathbf{x}'' &= \Big(u,\;v+\tfrac{f_y\,\delta}{D(\mathbf{x})}\Big),\\
\mathbf{f}^{y}(\mathbf{x}) &= \mathbf{x}''-\mathbf{x}
= \Big(0,\;\tfrac{f_y\,\delta}{D(\mathbf{x})}\Big).
\end{aligned}}
\end{equation}
{Its magnitude is also controlled by $1/D(\mathbf{x})$, capturing vertical parallax such as the vertical component of hand-held camera shake.}

\noindent\emph{{(c) Forward translation ($Z_{\text{cam}}$-axis).}} {For $\mathbf{P}'''=(X,Y,Z\!+\!\delta)$, the point moves along the viewing direction, so its depth changes. Re-projecting $\mathbf{P}'''$ scales the offset from the principal point, giving}
\begingroup
\small
\begin{equation}
{\begin{gathered}
\mathbf{x}''' =
\Big(c_x+\tfrac{D(\mathbf{x})}{D(\mathbf{x})+\delta}(u-c_x),\;
c_y+\tfrac{D(\mathbf{x})}{D(\mathbf{x})+\delta}(v-c_y)\Big),\\
\mathbf{x}'''-\mathbf{x}
= -\frac{\delta}{D(\mathbf{x})+\delta}\,\big(u-c_x,\;v-c_y\big).
\end{gathered}}
\end{equation}
\endgroup
{Since \(\delta\) is small relative to the scene depth, we use the first-order approximation as the forward basis:}
\begin{equation}
{\mathbf{f}^{z}(\mathbf{x}) \approx -\frac{\delta}{D(\mathbf{x})}\,\big(u-c_x,\;v-c_y\big).}
\end{equation}
{This basis is \emph{radial} around the principal point $(c_x,c_y)$. When the camera moves forward ($\delta\!<\!0$), pixels move outward; when it moves backward ($\delta\!>\!0$), pixels move inward. The magnitude decreases with depth, matching the familiar looming effect of forward camera motion. Under this first-order form, the global scale and sign of \(\delta\) are absorbed by the learned basis weight.}

{In summary, one depth map produces three analytic translation bases $\{\mathbf{f}^{x},\mathbf{f}^{y},\mathbf{f}^{z}\}$ for sideways, vertical, and forward camera motion. They are determined by the depth map $D(\mathbf{x})$ and intrinsics $\mathbf{K}$, while the remaining global scale and sign are absorbed by the learned basis weights. The key difference from homography bases is that these bases are depth-aware: their magnitudes change with scene depth and therefore explicitly encode translation-induced parallax.}

\begin{figure}[t]
    \centering
    \includegraphics[width=1\linewidth]{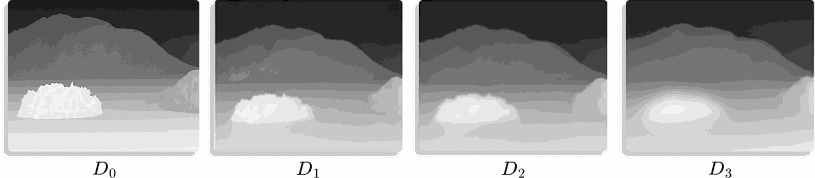}
    \caption{{Gaussian-smoothed depth pyramid. Starting from the original depth map $D_0$, progressively larger Gaussian kernels generate smoother maps $D_1$, $D_2$, and $D_3$. Fine levels preserve sharp foreground boundaries and background depth layers, while coarse levels suppress local details and emphasize smoother regional depth structure.}}
    \label{fig:depth_pyramid}
\end{figure}

{Using only the raw depth map can make the bases overly sensitive to local depth noise and sharp object boundaries. To make the representation more stable, we construct a Gaussian-smoothed depth pyramid $\mathcal{D}=\{D_s\}_{s=0}^{S-1}$ from the original depth map $D_0$:}
\begin{equation}
{D_s = G_{\sigma_s} * D_0,\qquad 0=\sigma_0<\sigma_1<\cdots<\sigma_{S-1},}
\end{equation}
{where $G_{\sigma_s}$ is a Gaussian kernel and $*$ denotes convolution. As illustrated in Fig.~\ref{fig:depth_pyramid}, increasing the kernel size from $D_1$ to $D_3$ progressively smooths the depth map. In $D_0$, the foreground object has sharp boundaries and the background mountains retain visible layered depth contours. After smoothing, these fine structures gradually fade: the foreground object becomes more diffuse, the mountain contours merge into broader depth regions, and the transitions between the foreground, water surface, and background become softer. Thus, $D_0$ keeps object-level parallax details, while coarser maps such as $D_3$ emphasize low-frequency regional depth structure. For each pyramid level $D_s$, we generate the three translation bases $\{\mathbf{f}^{x}_{s},\mathbf{f}^{y}_{s},\mathbf{f}^{z}_{s}\}$ using the same analytic formulas above. The full depth-translational basis set is therefore}
\begin{equation}
{\mathcal{B}_{\mathrm{trans}}=\big\{\mathbf{f}^{x}_{s},\mathbf{f}^{y}_{s},\mathbf{f}^{z}_{s}\big\}_{s=0}^{S-1}.}
\end{equation}
{Each depth-translational basis can also be transformed into an optical flow according to the image coordinate, as shown in Fig.~\ref{fig:network-ppl} (2nd row of Physical motion bases). Together, the homography bases and depth-translational bases form our physical basis family, covering both homography-like motion and depth-induced parallax.}

\subsubsection{Stochastic Basis.} While the physical basis family captures explicit geometric motion patterns, the complete space of camera motions is still infinitely dimensional, and residual composite motion cannot be exhaustively enumerated. Inspired by stochastic modeling strategies in image restoration~\cite{wang2022zero}, where random components complement deterministic reconstruction to recover richer residual details, we introduce stochastic bases to complement the explicit physical bases.
{These bases capture residual composite motion that cannot be conveniently expressed by analytic geometric construction alone.} Specifically, we generate $N_s$ random $3\!\times\!3$ matrices \(\left\{ \mathbf{H}^{(k)} \right\}_{k=1}^{N_s} \subseteq \mathbb{R}^{3 \times 3} \), where each matrix is formulated as:
\begingroup
\small
\begin{equation}
\mathbf{H} = \{ h_i \}_{i=1}^9, \text{where} \quad
h_i = \begin{cases}
\epsilon_i \sim \mathcal{N}(0,1), & 1 \le i \le 8, \\
1, & i = 9.
\end{cases}
\end{equation}
\endgroup

Following BasesHomo~\cite{BasesHomo2021}, we convert random matrices into flows and apply singular value decomposition (SVD) to extract principal components. This process yields stochastic bases that capture diverse motion patterns beyond the explicit geometric basis families, as illustrated in Fig.~\ref{fig:network-ppl} (labeled as stochastic motion bases). {Together with the physical basis family, they form a hybrid basis set that balances geometric interpretability with the flexibility required to approximate richer composite camera motions.}

\subsection{Network Structure}
\label{subsec:HEM}


The network is illustrated in Fig.~\ref{fig:network-ppl}. Given an input frame pair, we follow practices from previous work~\cite{HomoGAN2022,CA-Unsupervised2020} and first crop it into $320\! \times\! 576$ patches. We denote the cropped reference and target patches as \(I_a\) and \(I_b\), respectively. The patches are then converted to grayscale, projected into shallow feature maps \(f_a\) and \(f_b\) to handle luminance variations, and processed by a $3$-layer feature pyramid for multi-scale motion estimation. {Then we propose a Motion Estimation Transformer (MET). MET takes the paired multi-scale feature pyramids as input and follows the token-block design of HomoGAN~\cite{HomoGAN2022}, but changes the output head from 8 basis weights to \(N\)-dimensional basis coefficients for each direction. The MET outputs one global coefficient vector for each direction, \(\hat{\mathbf{w}}_{ab},\hat{\mathbf{w}}_{ba}\in\mathbb{R}^{N}\). Since depth-translational bases are defined in the coordinate domain of each reference frame, we construct direction-specific hybrid basis sets \(\mathcal{B}_a\) and \(\mathcal{B}_b\) using the corresponding cropped depth maps \(D_a\) and \(D_b\), respectively. Before linear combination, all basis functions are rasterized into pixel-displacement flow maps on the cropped image grid, so homography, depth-translational, and stochastic bases share the same coordinate and scale. Instead of directly regressing a dense motion field, the MET predicts combination weights over these hybrid basis sets, which contain homography-based bases, depth-translational bases, and stochastic bases.}
{The final camera motion is then computed through a linear combination of the predicted weights and their corresponding motion bases. The bidirectional dense motion fields are}
\begin{equation}
{\hat{\mathbf{m}}_{ab}=\sum_{\mathbf{b}_{a,k}\in\mathcal{B}_{a}}\hat{w}_{ab}^{k}\mathbf{b}_{a,k},\qquad
\hat{\mathbf{m}}_{ba}=\sum_{\mathbf{b}_{b,k}\in\mathcal{B}_{b}}\hat{w}_{ba}^{k}\mathbf{b}_{b,k}.}
\end{equation}
{and the subset assigned to the depth-translational bases forms the translation-flow component:}
\begin{equation}
{\hat{\mathbf{t}}_{ab}=\sum_{\mathbf{b}_{a,k}\in\mathcal{B}_{\mathrm{trans}}^{a}}\hat{w}_{ab}^{k}\mathbf{b}_{a,k},\qquad
\hat{\mathbf{t}}_{ba}=\sum_{\mathbf{b}_{b,k}\in\mathcal{B}_{\mathrm{trans}}^{b}}\hat{w}_{ba}^{k}\mathbf{b}_{b,k},}
\end{equation}
{where $\mathcal{B}_{\mathrm{trans}}^{a}$ and $\mathcal{B}_{\mathrm{trans}}^{b}$ denote the depth-translational basis subsets constructed from \(D_a\) and \(D_b\), respectively. We use the dense flows to warp both images and feature maps. Specifically, the forward flow \(\hat{\mathbf{m}}_{ab}\) warps the target image and feature map toward the reference frame, \(I_b'=\mathcal{W}(I_b,\hat{\mathbf{m}}_{ab})\) and \(f_b'=\mathcal{W}(f_b,\hat{\mathbf{m}}_{ab})\), while the backward flow \(\hat{\mathbf{m}}_{ba}\) warps the reference image and feature map toward the target frame, \(I_a'=\mathcal{W}(I_a,\hat{\mathbf{m}}_{ba})\) and \(f_a'=\mathcal{W}(f_a,\hat{\mathbf{m}}_{ba})\). The warped feature maps are used for feature consistency and confidence-map prediction.}
The confidence maps $\mathbf{d}_{ab}$ and $\mathbf{d}_{ba}$ are crucial for training robust camera-motion estimation, because they down-weight dynamic objects and poorly aligned regions in the loss. Specifically, the mask network $\mathcal{M}$~\cite{GyroFlow} takes the original and warped feature maps as input and predicts confidence maps that highlight well-aligned regions, i.e., \(\mathbf{d}_{ab}=\mathcal{M}([f_a,f_b'])\) and \(\mathbf{d}_{ba}=\mathcal{M}([f_b,f_a'])\).

\subsection{Loss Function}
\label{subsec:loss}

\noindent\textbf{Probabilistic Motion Modeling.} Camera motion estimation faces a fundamental challenge: distinguishing between camera-induced and object motion in the scene. We therefore formulate our approach as a probabilistic model that explicitly accounts for uncertainty in motion estimation.

Building on previous findings that 2D motion follows a Laplace distribution~\cite{ilg2018uncertainty, gast2018lightweight, PDC-Net}, we model the conditional probability of the available camera-motion supervision given our prediction:
\begin{equation}
p(\mathbf{m}_{ab} \mid \hat{\mathbf{m}}_{ab};\mathbf{d}_{ab}),
\end{equation}
where $\hat{\mathbf{m}}_{ab}$ is our predicted motion field and $\mathbf{d}_{ab}$ represents our confidence in each pixel's motion estimate. Higher confidence values indicate pixels likely following camera motion, while lower values suggest pixels belonging to independently moving objects.

\noindent\textbf{Laplace Distribution Model.} We use a confidence-weighted Laplace negative log-likelihood for both motion and feature supervision. For a target field \(\mathbf{y}\), prediction \(\hat{\mathbf{y}}\), and confidence map \(\mathbf{d}\), we define
\begingroup
\small
\begin{equation}
\label{eq:nll}
{\begin{aligned}
\ell_{\mathrm{Lap}}(\mathbf{y},\hat{\mathbf{y}};\mathbf{d})
&=\mathbb{E}_{\mathbf{x}}\sum_{c}\Big[
\sqrt{2(\mathbf{d}(\mathbf{x})+\epsilon)}
\left|\mathbf{y}^{c}(\mathbf{x})-\hat{\mathbf{y}}^{c}(\mathbf{x})\right|\\
&\qquad\qquad\qquad\qquad
-\frac{1}{2}\log(\mathbf{d}(\mathbf{x})+\epsilon)\Big].
\end{aligned}}
\end{equation}
\endgroup
Here \(c\) indexes channels: for motion, \(c\in\{u,v\}\) denotes the horizontal and vertical components; for features, \(c\) indexes feature channels. This is equivalent to setting the per-pixel variance to \(\sigma^2(\mathbf{x})=1/(\mathbf{d}(\mathbf{x})+\epsilon)\), so high confidence corresponds to small variance and stronger supervision. Since low confidence corresponds to large variance, the data term alone could be reduced by assigning low confidence to difficult pixels. The log-normalizer penalizes overly large variance, preventing the trivial solution of assigning low confidence to all pixels.

\noindent\textbf{Hybrid Loss Strategy.} Another challenge in camera motion estimation is the scarcity of ground-truth labels. Therefore, we introduce a hybrid loss strategy with three components:
1) {Motion supervision loss} ($\ell_{\mathit{NLL}_m}$): we use homography-derived pseudo labels and apply negative log-likelihood (NLL) loss in both forward and backward directions:
\begingroup
\small
\begin{equation}
{\ell_{\mathit{NLL}_m}=\ell_{\mathrm{Lap}}(\mathbf{m}_{ab},\hat{\mathbf{m}}_{ab};\mathbf{d}_{ab})+\ell_{\mathrm{Lap}}(\mathbf{m}_{ba},\hat{\mathbf{m}}_{ba};\mathbf{d}_{ba}).}
\end{equation}
\endgroup

\noindent2) {Feature consistency loss} ($\ell_{\mathit{NLL}_{f}}$): We apply the same probabilistic framework to enforce consistency between warped features. Given image features $f_a$ and $f_b$, we use our predicted motion to produce warped features $f_a'$ and $f_b'$:
\begingroup
\small
\begin{equation}
{\ell_{\mathit{NLL}_{f}}=\ell_{\mathrm{Lap}}(f_a,f_b';\mathbf{d}_{ab})+\ell_{\mathrm{Lap}}(f_b,f_a';\mathbf{d}_{ba}).}
\end{equation}
\endgroup

\noindent3) {Depth-aware smoothness loss} ($\ell_{\mathrm{smooth}}$): We regularize the translation-induced flows $\hat{\mathbf{t}}_{ab}$ and $\hat{\mathbf{t}}_{ba}$ to be smooth within continuous-depth regions while allowing changes near depth discontinuities:
\begingroup
\small
\begin{equation}
{\begin{aligned}
\ell_{\mathrm{smooth}}=\mathbb{E}_{\mathbf{x}}\!\Big[
&e^{-\alpha\|\nabla D_a(\mathbf{x})\|_1}\|\nabla \hat{\mathbf{t}}_{ab}(\mathbf{x})\|_1\\
&+e^{-\alpha\|\nabla D_b(\mathbf{x})\|_1}\|\nabla \hat{\mathbf{t}}_{ba}(\mathbf{x})\|_1
\Big].
\end{aligned}}
\end{equation}
\endgroup
{Here \(\mathbb{E}_{\mathbf{x}}\) denotes averaging over image pixels, \(\nabla\) is the spatial image-gradient operator, \(D_a\) and \(D_b\) are the depth maps of the two frames, and $\alpha$ controls the sensitivity to depth edges.}
The overall loss is:
\begingroup
\small
\begin{equation}
\label{eq:overall_loss}
\ell_{\mathit{overall}} = \ell_{\mathit{NLL}_{m}} + \lambda_f\ell_{\mathit{NLL}_{f}} + \lambda_s\ell_{\mathrm{smooth}},
\end{equation}
\endgroup
where $\lambda_f$ and $\lambda_s$ are predefined weights.

\section{Experiments}
\label{subsec:experiment}

\subsection{Dataset}
\label{subsec:dataset}
{We use CAHomo~\cite{CA-Unsupervised2020} and GHOF~\cite{gyroflow+} as existing benchmarks, and use our previously introduced GHOF-Cam benchmark for dense camera-motion evaluation. Unless otherwise specified, models are trained on CAHomo with additional generated samples~\cite{li2024dmhomo}, evaluated zero-shot on GHOF/GHOF-Cam, and further refined on the GHOF training split when studying target-domain adaptation.}

{Existing evaluation protocols leave a gap between sparse background camera-motion evaluation and dense generic optical-flow estimation. GHOF provides both dense optical-flow annotations and sparse homography correspondences for camera-motion evaluation. However, its dense optical-flow labels include independently moving objects and occlusion regions, which are not the target signal for camera-motion estimation. To obtain dense camera-motion evaluation, the conference version constructed GHOF-Cam from GHOF by retaining dense flow only on valid static regions. Thus, GHOF is used for sparse/background camera-motion evaluation, while GHOF-Cam extends the evaluation to dense valid static pixels.}

\begin{figure}
    \centering
    \includegraphics[width=1\linewidth]{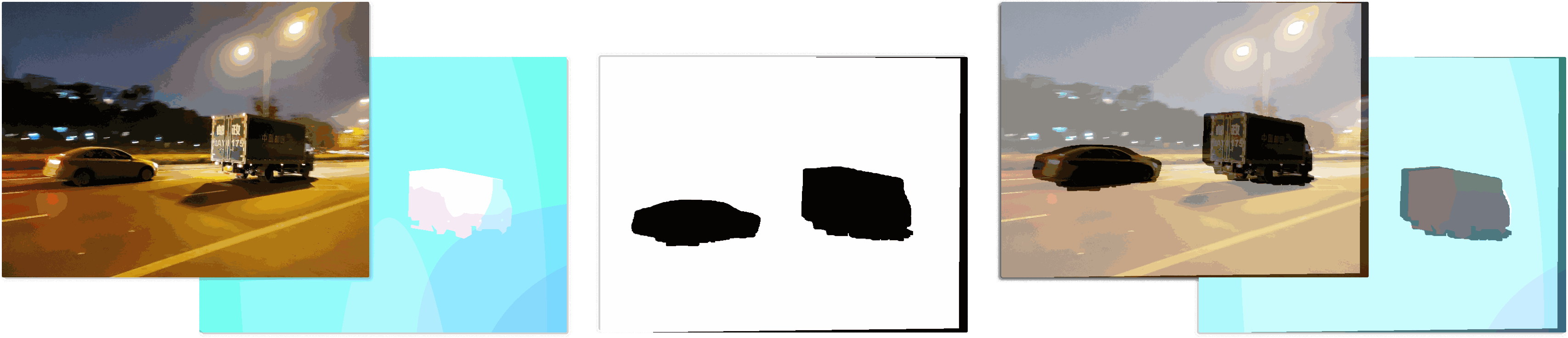}
    \caption{Construction of GHOF-Cam. We mask dynamic objects in GHOF using SAM-assisted annotations, dilate the masks to cover occlusion boundaries, and remove invalid edge regions from homography-induced black borders. The remaining static pixels provide dense camera-motion labels for evaluation.}
    \label{fig:GHOF-Mask}
    \vspace{-1em}
\end{figure}

{\noindent\textbf{GHOF-Cam Benchmark.} GHOF-Cam, introduced in the conference version, evaluates dense camera motion under realistic mobile-camera conditions. As shown in Fig.~\ref{fig:GHOF-Mask}, we first employ the Segment Anything Model (SAM)~\cite{kirillov2023segment} to produce segmentation candidates and then identify dynamic categories such as cars and people. The resulting object masks are dilated to include object-boundary occlusions, where optical flow is often ill-posed. Since semantic masks do not cover all invalid pixels, we further use the ground-truth homography to detect black edge regions caused by image warping and add them to the invalid mask. We finally apply the combined masks to both input images and ground-truth optical flow. The remaining pixels provide dense supervision/evaluation targets for static-scene motion induced by the camera, making GHOF-Cam complementary to CAHomo's sparse homography protocol and GHOF's original sparse/background or generic optical-flow protocols.}

\subsection{Implementation Details}
\label{subsec:Implementation}
CamFlow builds on HomoGAN~\cite{HomoGAN2022}. We keep its feature pyramid and token-block design, but replace the original 8-basis output head with an $N$-dimensional motion-basis head. Unless otherwise stated, CamFlow uses 24 bases: 12 homography-derived physical bases and 12 stochastic bases extracted from randomly sampled homography flows by SVD. CamFlow+ further adds 12 depth-translational bases, generated from four depth-pyramid levels with three translation directions at each level.

For the CAHomo pretraining used by the 8- and 12-basis physical-basis variants in Sec.~\ref{sec:abla_study}, we use the Adam optimizer~\cite{kingma2014adam} with learning rate $1.0\times10^{-5}$, $\beta_1=0.9$, and $\beta_2=0.99$. The learning rate decays by 0.8 after each epoch. We train for 5 epochs with batch size 16, using $600\times800$ image pairs cropped to $320\times576$ patches. To isolate the effect of the physical motion bases, the weight of $\ell_{\mathit{NLL}_{m}}$ is 1, while the other loss weights are set to 0 in this pretraining setting. Training takes about 14 hours on 4 NVIDIA A800 GPUs. Unless a result is explicitly marked as this physical-basis pretraining setting, the shared 24-basis configuration uses the hybrid probabilistic loss ablated in Sec.~\ref{sec:abla_study}.

For GHOF refinement, we use the same optimization setting on the GHOF training split, with the loss weights for $\ell_{\mathit{NLL}_{m}}$, $\ell_{\mathit{NLL}_{f}}$, and $\ell_{\mathrm{smooth}}$ set to $1:10:40$. CamFlow+ requires camera intrinsics and depth to construct depth-translational bases. GHOF provides intrinsics, and we estimate per-frame depth using VGGT~\cite{wang2025vggt}; the estimated depth maps are cropped consistently with the input image patches and then used to build the depth pyramid. Since CAHomo does not provide intrinsics, CamFlow+ is trained only in the GHOF refinement setting. For fair comparison, we also refine the strongest trainable baselines on the same GHOF split, while traditional methods are used without training.

\begin{table}[t]
\caption{The benchmark consists of 5 distinct scenarios, namely regular (RE), low-texture (LT), low-light (LL), small foreground (SF), and large foreground (LF). The point matching errors (PME) on the test set of CAHomo~\cite{CA-Unsupervised2020} are presented. For each column, the best result is highlighted in \textbf{bold} and the second-best is \underline{underlined}.}
\label{tab:Compare_CAunsup}
\vspace{-2.5em}
  \begin{center}
  \resizebox{1\linewidth}{!}{
  \begin{tabular}{r|l|lllllll}
  \toprule
  & Methods & \multicolumn{1}{c}{AVG} & \multicolumn{1}{c}{RE} & \multicolumn{1}{c}{LT} & \multicolumn{1}{c}{LL} & \multicolumn{1}{c}{SF} & \multicolumn{1}{c}{LF} \\
  \midrule
  1) & $\mathcal{I}_{3\times3}$ & 6.70  & 7.75  & 7.65  & 7.21  & 7.53  & 3.39  \\
  \midrule
  2) & SIFT\cite{sift} + RANSAC\cite{ransac} & 1.41  & 0.30  & 1.34  & 4.03  & 0.81  & 0.57  \\
  3) & SIFT\cite{sift} + MAGSAC\cite{magsac} & 1.34  & 0.31  & 1.72  & 3.39  & 0.80  & 0.47  \\
  4) & ORB\cite{orb} + RANSAC\cite{ransac} & 1.48  & 0.85  & 2.59  & 1.67  & 1.10  & 1.24  \\
  5) & ORB\cite{orb} + MAGSAC\cite{magsac} & 1.69  & 0.97  & 3.34  & 1.58  & 1.15  & 1.40  \\
  6) & SPSG\cite{superpoint,superglue} + RANSAC\cite{ransac} & 0.71  & 0.41  & 0.87  & 0.72  & 0.80  & 0.75  \\
  7) & SPSG\cite{superpoint,superglue} + MAGSAC\cite{magsac} & 0.63  & 0.36  & 0.79  & 0.70  & 0.71  & 0.70  \\
  8) & LoFTR\cite{loftr} + RANSAC\cite{ransac} & 1.44  & 0.56  & 2.70  & 1.36  & 1.05  & 1.52  \\
  9) & LoFTR\cite{loftr} + MAGSAC\cite{magsac} & 1.39  & 0.55  & 2.57  & 1.33  & 1.05  & 1.41  \\
  \midrule

  10) & DHN\cite{supervised2016} & 2.87  & 1.51  & 4.48  & 2.76  & 2.62  & 3.00  \\
  11) & LocalTrans\cite{Crossresolution-supervised2021} & 4.21  & 4.09  & 4.84  & 4.55  & 5.30  & 2.25  \\
  12) & IHN\cite{Iterative-supervised2022} & 4.67  & 4.85  & 5.54  & 5.10  & 5.04  & 2.84  \\

  13) & {$\textrm{RealSH}$\cite{jiang2023supervised}} & {0.34 } & {0.22 } & {\underline{0.35} } & {{0.44} } & {0.42 } & {\underline{0.29} }\\
  14) & {$\textrm{DMHomo}$\cite{li2024dmhomo}} & {\textbf{0.31} } & {\textbf{0.19} } & {\textbf{0.33} } & {\textbf{0.40} } & {\textbf{0.38} } & {\textbf{0.28} } \\
  \midrule

  15) & CAHomo\cite{CA-Unsupervised2020} & 0.88  & 0.73  & 1.01  & 1.03  & 0.92  & 0.70  \\
  16) & BasesHomo\cite{BasesHomo2021} & 0.50  & 0.29  & 0.54  & 0.65  & 0.61  & 0.41  \\
  17) & HomoGAN\cite{HomoGAN2022} & {0.39 } & {0.22 } & {0.41 } & 0.57  & {0.44 } & {0.31 }\\
  \midrule
  18) & {CamFlow} & {\underline{0.33} } & {\underline{0.20} } & {\textbf{0.33} } & {\underline{0.41} } & {\underline{0.39} } & {0.30 } \\
   \bottomrule
   \end{tabular}}
\end{center}
\end{table}

\begin{table}[t]
\caption{Zero-shot and fine-tuned point matching errors (PME) on the GHOF~\cite{gyroflow+} test set. Methods marked with $*$ are fine-tuned on the GHOF training split. For each column, the best result is highlighted in \textbf{bold} and the second-best is \underline{underlined}.}
\label{tab:Compare_GHOF}
\vspace{-2.5em}
  \begin{center}
  \resizebox{1\linewidth}{!}{
  \begin{tabular}{rl|ccccccc}
  \toprule
  1) & Methods & AVG & RE & FOG & LL & RAIN & SNOW \\
    \midrule
    2) & $\mathcal{I}_{3\times3}$ & 6.33  & 4.94  & 7.24  & 8.09  & 5.48  & 5.89  \\
      \midrule
    3) & SIFT\cite{sift} & 4.80 & {0.59}  & 4.47  & 12.10 & 0.62  & 6.20   \\
    4) & {SOSNet\cite{sosnet}} & {5.27} & {0.72} & {5.56} & {12.95} & {0.77} & {6.37} \\
    5) & SPSG\cite{superpoint,superglue} & 4.47 & 3.54  & 2.21  & 10.66 & 0.83  & 5.10   \\
    6) & {LoFTR\cite{loftr}} & {3.04} & {2.64} & {1.33} & {6.71} & {0.56} & {3.94} \\
    7) & {HM-Mix\cite{grundmann2012calibration}} & {4.15} & {3.42} & {2.35} & {6.41} & {3.78} & {4.80} \\

    \midrule
    8) & DHN\cite{supervised2016} & 6.61  & 6.04  & 6.02  & 7.68  & 6.99  & 6.32  \\
    9) & LocalTrans\cite{Crossresolution-supervised2021} & 5.72 & 4.06 & 6.49 & 5.95 & 5.78 & 6.34 \\
    10) & IHN\cite{Iterative-supervised2022} & 8.17 & 7.10 & 8.71 & 9.34 & 6.57 & 9.13 \\
    11) & RealSH\cite{jiang2023supervised} & {1.72} & 1.60 & 0.88 & 4.42 & {0.43} & {1.28} \\
    12) & DMHomo\cite{li2024dmhomo} & 1.75 & {0.64} & 0.85 & 4.16 & \underline{0.39} & 2.74 \\
    \midrule
    13) & CAHomo\cite{CA-Unsupervised2020} & 3.87  & 4.10  & 3.84 & 6.99  & 1.27  & 3.17  \\
    14) & BasesHomo\cite{BasesHomo2021} & 2.28  & 2.02  & 1.43  & 4.90  & 0.78  & 2.29  \\
    15) & HomoGAN\cite{HomoGAN2022} & 1.95 & 1.73 & {0.60} & {3.95} & 0.47 & 3.02 \\
    16) & CamFlow & {1.23} & 1.15 & {0.96} & {2.69} & {0.40} & {0.93} \\
    \midrule
    17) & {VGGT\cite{wang2025vggt}} & {1.96} & {1.55} & {1.07} & {3.31} & {1.80} & {2.09} \\
    18) & {DUSt3R\cite{dust3r_cvpr24}} & {5.56} & {4.65} & {6.28} & {7.52} & {4.46} & {4.91} \\
    19) & {RAFT\cite{teed2020raft}} & {1.87} & {2.83} & {0.77} & {3.04} & {\underline{0.39}} & {2.32} \\
    20) & {Sea-RAFT\cite{wang2024sea}} & {1.74} & {2.56} & {0.73} & {3.52} & {0.61} & {1.29} \\
    \midrule
    21) & {HomoGAN*\cite{HomoGAN2022}} & {0.79} & {0.75} & {0.52} & {1.39} & {0.59} & {0.69} \\
    22) & {DMHomo*\cite{li2024dmhomo}} & {0.70} & {0.60} & {0.47} & {1.35} & {0.47} & {\underline{0.63}} \\
    23) & {CamFlow*} & {\underline{0.64}} & {\underline{0.55}} & {\underline{0.39}} & {\underline{1.15}} & {0.41} & {0.68} \\
    24) & {CamFlow+} & {\textbf{0.51}} & {\textbf{0.31}} & {\textbf{0.28}} & {\textbf{0.99}} & {\textbf{0.37}} & {\textbf{0.58}} \\
  \bottomrule
  \end{tabular}}
  \end{center}
\end{table}%

\begin{table}[t]
\caption{Zero-shot and fine-tuned end point errors (EPE) on GHOF-Cam. Methods marked with $*$ are fine-tuned on the GHOF training split. For each column, the best result is highlighted in \textbf{bold} and the second-best is \underline{underlined}.}
\label{tab:GHOF-Cam-Quan}
\vspace{-2.5em}
  \begin{center}
  \resizebox{1\linewidth}{!}{
  \begin{tabular}{rl|ccccccc}
  \toprule
  1) & Methods & AVG & RE & FOG & LL & RAIN & SNOW \\
    \midrule
    2) & $\mathcal{I}_{3\times3}$ &5.22 &3.65 & 6.69 & 5.88  & 4.90  & 4.96  \\
    \midrule
    3) & SIFT\cite{sift} &2.82 &0.60 & 2.43 & 7.09  & 0.61  & 3.37  \\
    4) & SPSG\cite{superpoint,superglue} &3.07 &3.99 & 1.57 & 6.88  &0.79  & 2.16  \\
    5) & MeshFlow\cite{LiuTYSZ16} &2.15 &1.09&2.21&5.57&0.44 &1.69 \\
    6) & HM-Mix\cite{grundmann2012calibration} &4.35 &1.02 & 4.03 & 8.75  & 1.53  & 6.42  \\
    7) & RANSAC-F\cite{shen2020ransac} &3.26 &2.81 &3.14  &5.12  & 2.21  &3.04   \\
    \midrule
    8) & CAHomo\cite{CA-Unsupervised2020} &2.81 &2.02 & 2.03 & 4.56  & 2.84  & 2.61  \\
    9) & BasesHomo\cite{BasesHomo2021} &1.74 &1.39 & 0.97 & 4.12  & 0.66  & 1.58  \\
    10) & {HomoGAN\cite{HomoGAN2022}} & {1.16} & {1.61} & {0.47} & {1.94}  & {0.45}  & {1.35} \\
    11) & CamFlow &1.10 &1.08 & 0.74 & 2.15  & 0.46  & 1.05  \\
    \midrule

    12) & {VGGT\cite{wang2025vggt}} & {1.88} & {1.24} & {1.08} & {3.63} & {1.64} & {1.67} \\
    13) & {DUSt3R\cite{dust3r_cvpr24}} & {4.68} & {3.18} & {5.39} & {6.59} & {3.97} & {4.29} \\
    14) & {RAFT\cite{teed2020raft}} & {0.89} & {\underline{0.36}} & {0.42} & {2.12} & {\textbf{0.32}} & {1.24} \\
    15) & {Sea-RAFT\cite{wang2024sea}} & {0.89} & {\textbf{0.24}} & {0.55} & {2.59} & {\underline{0.39}} & {0.68} \\
    \midrule

    16) & {HomoGAN*\cite{HomoGAN2022}} & {0.73} & {0.81} & {0.42} & {1.16}  & {0.60}  & {0.68} \\
    17) & {DMHomo*\cite{li2024dmhomo}} & {0.70} & {0.81} & {0.40} & {1.13}  & {0.49}  & {\underline{0.65}} \\
    18) & {CamFlow*} & {\underline{0.64}} & {0.71} & {\underline{0.31}} & {\underline{1.04}} & {0.45} & {0.71} \\
    19) & {CamFlow+} & {\textbf{0.50}} & {0.42} & {\textbf{0.21}} & {\textbf{0.86}}  & {0.40}  & {\textbf{0.59}}  \\
  \bottomrule
  \end{tabular}}
  \end{center}
\end{table}%

\begin{figure}[th]
\begin{center}
  \includegraphics[width=1\linewidth]{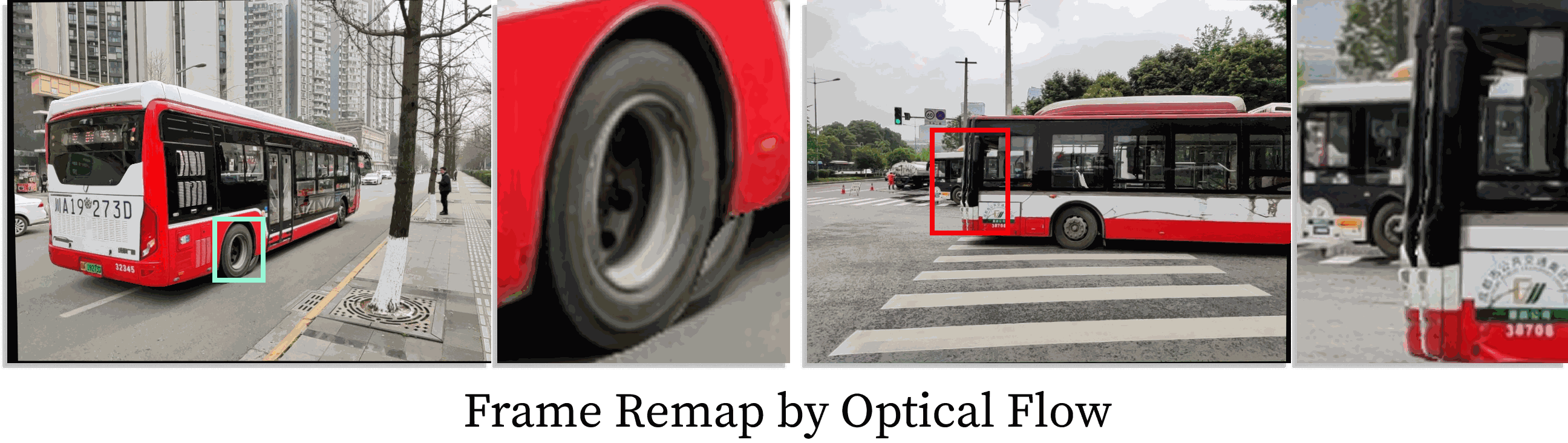}
\end{center}
  \caption{{Image transformation with optical flow can introduce ghosting and distortion around occlusion and dynamic-object regions, highlighting the need to isolate camera-induced background motion for downstream video enhancement tasks.}}
\label{fig:of_remap}
\end{figure}

\begin{table*}[ht]
\caption{Perceptual quality comparison on GHOF-Cam under different conditions. We report PSNR and SSIM (higher is better) and LPIPS (lower is better), computed within the valid static masks. For each metric column, the best result is highlighted in \textbf{bold} and the second-best is \underline{underlined}; the identity baseline $\mathcal{I}_{3\times3}$ and the GT-Homo oracle are excluded from the ranking.}
\label{tab:ghof-cam}
\vspace{-1em}
\centering
\resizebox{\linewidth}{!}{%
\begin{tabular}{l|ccc|ccc|ccc|ccc|ccc|ccc}
\toprule
\multirow{2}{*}{Method} &
\multicolumn{3}{c|}{AVG} &
\multicolumn{3}{c|}{RE} &
\multicolumn{3}{c|}{FOG} &
\multicolumn{3}{c|}{LL} &
\multicolumn{3}{c|}{RAIN} &
\multicolumn{3}{c}{SNOW} \\
\cmidrule(lr){2-4} \cmidrule(lr){5-7} \cmidrule(lr){8-10} \cmidrule(lr){11-13} \cmidrule(lr){14-16} \cmidrule(lr){17-19}
 & PSNR$\uparrow$ & SSIM$\uparrow$ & LPIPS$\downarrow$ & PSNR$\uparrow$ & SSIM$\uparrow$ & LPIPS$\downarrow$ & PSNR$\uparrow$ & SSIM$\uparrow$ & LPIPS$\downarrow$ & PSNR$\uparrow$ & SSIM$\uparrow$ & LPIPS$\downarrow$ & PSNR$\uparrow$ & SSIM$\uparrow$ & LPIPS$\downarrow$ & PSNR$\uparrow$ & SSIM$\uparrow$ & LPIPS$\downarrow$ \\
\midrule
$\mathcal{I}_{3\times3}$ & 24.05 & 0.7403 & 0.0836 & 21.06 & 0.6900 & 0.0750 & 26.57 & 0.7711 & 0.0821 & 25.70 & 0.8506 & 0.0785 & 21.53 & 0.5335 & 0.1411 & 25.37 & 0.8562 & 0.0412 \\
GT-Homo & 32.78 & 0.9187 & 0.0570 & 28.39 & 0.8697 & 0.0549 & 35.23 & 0.9508 & 0.0492 & 31.88 & 0.9405 & 0.0575 & 30.11 & 0.8511 & 0.1033 & 38.31 & 0.9814 & 0.0199 \\
\midrule
SIFT & 28.44 & 0.9074 & 0.0781 & \textbf{29.23} & 0.9148 & 0.0545 & 29.42 & 0.9016 & 0.0768 & 27.37 & 0.9074 & 0.0982 & 30.00 & 0.8632 & 0.1055 & 26.16 & 0.9497 & 0.0558 \\
SPSG & 28.01 & 0.8697 & 0.0796 & 21.83 & 0.7593 & 0.0886 & 30.88 & 0.9049 & 0.0645 & 27.60 & 0.9019 & 0.0966 & 28.86 & 0.8270 & 0.1103 & 30.88 & 0.9556 & 0.0379 \\
MeshFlow & 29.91 & 0.9239 & 0.0688 & 28.57 & \textbf{0.9216} & 0.0576 & 28.68 & 0.9280 & 0.0742 & 29.41 & 0.9254 & 0.0774 & \textbf{30.68} & \textbf{0.8747} & 0.1049 & 32.23 & 0.9700 & 0.0298 \\
HM-Mix & 25.77 & 0.8896 & 0.0882 & 26.09 & 0.8721 & 0.0596 & 26.56 & 0.8753 & 0.0882 & 26.43 & 0.9037 & 0.1002 & 28.20 & 0.8672 & 0.1107 & 21.58 & 0.9296 & 0.0820 \\
RANSAC-F & 26.04 & 0.8348 & 0.0890 & 26.09 & 0.8812 & 0.0665 & 29.22 & 0.8944 & 0.0801 & 27.29 & 0.9031 & 0.0923 & 21.68 & 0.5585 & 0.1495 & 25.90 & 0.9371 & 0.0566 \\
CAHomo & 25.29 & 0.7837 & 0.0841 & 22.67 & 0.7341 & 0.0805 & 27.51 & 0.8048 & 0.0751 & 26.12 & 0.8743 & 0.0846 & 22.95 & 0.6130 & 0.1420 & 27.20 & 0.8924 & 0.0384 \\
BasesHomo & 29.61 & 0.9026 & 0.0672 & 25.08 & 0.8522 & 0.0666 & 31.06 & 0.9170 & 0.0627 & 30.05 & 0.9303 & 0.0702 & 29.58 & 0.8512 & 0.1071 & 32.30 & 0.9622 & 0.0292 \\
{VGGT} & {28.81} & {0.8692} & {0.0577} & {25.14} & {0.8432} & {0.0536} & {31.79} & {0.9065} & {0.0512} & {30.81} & {0.9305} & {0.0506} & {26.44} & {0.7240} & {0.1122} & {29.83} & {0.9294} & {0.0253} \\

\midrule
{HomoGAN} & {30.10} & {0.9176} & {0.0656} & {26.85} & {0.8667} & {0.0597} & {33.12} & {\textbf{0.9511}} & {0.0558} & {30.81} & {0.9391} & {0.0663} & {30.24} & {0.8645} & {0.1051} & {29.48} & {0.9666} & {0.0412} \\
{DMHomo} & {31.68} & {0.9218} & {0.0505} & {26.07} & {0.8789} & {0.0480} & {\underline{34.38}} & {0.9464} & {\underline{0.0446}} & {\underline{32.22}} & {\underline{0.9434}} & {\underline{0.0476}} & {30.03} & {0.8625} & {0.0972} & {35.92} & {\underline{0.9752}} & {\textbf{0.0183}} \\
{CamFlow} & {\underline{32.03}} & {\underline{0.9275}} & {\underline{0.0497}} & {26.64} & {0.8918} & {\underline{0.0458}} & {\textbf{34.67}} & {\underline{0.9505}} & {0.0448} & {32.14} & {\textbf{0.9435}} & {0.0478} & {30.33} & {\underline{0.8740}} & {\textbf{0.0949}} & {\textbf{36.60}} & {\textbf{0.9756}} & {\underline{0.0184}} \\
{CamFlow+} & {\textbf{32.30}} & {\textbf{0.9320}} & {\textbf{0.0487}} & {\underline{28.91}} & {\underline{0.9204}} & {\textbf{0.0413}} & {34.04} & {0.9486} & {\textbf{0.0444}} & {\textbf{32.24}} & {0.9432} & {\textbf{0.0473}} & {\underline{30.40}} & {0.8709} & {\underline{0.0953}} & {\underline{36.04}} & {0.9731} & {0.0187} \\
\bottomrule
\end{tabular}
}
\end{table*}

\subsection{Comparison with Existing Methods}
\label{subsec:compare_experiments}
We compare CamFlow and CamFlow+ with representative feature-based, learning-based, multi-plane, 3D foundation, and optical-flow methods. Feature methods include SIFT~\cite{sift}, ORB~\cite{orb}, SPSG~\cite{superpoint,superglue}, and LoFTR~\cite{loftr} with RANSAC~\cite{ransac} or MAGSAC~\cite{magsac}. Learning baselines include supervised methods~\cite{supervised2016,Crossresolution-supervised2021,Iterative-supervised2022,jiang2023supervised,li2024dmhomo} and unsupervised methods~\cite{CA-Unsupervised2020,BasesHomo2021,HomoGAN2022}. We also include MeshFlow~\cite{LiuTYSZ16}, Homography Mixture~\cite{grundmann2012calibration}, RANSAC-Flow~\cite{shen2020ransac}, DUSt3R~\cite{dust3r_cvpr24}, VGGT~\cite{wang2025vggt}, RAFT~\cite{teed2020raft}, and Sea-RAFT~\cite{wang2024sea}. DHN, LocalTrans, and IHN are pre-trained on MS-COCO~\cite{MSCOCO}; other learning-based homography methods are pre-trained on CAHomo unless otherwise stated.

\subsection{Quantitative Comparison.}
\label{subsec:quanti_compare}

\begin{figure*}[th]
\begin{center}
  \includegraphics[width=1\linewidth]{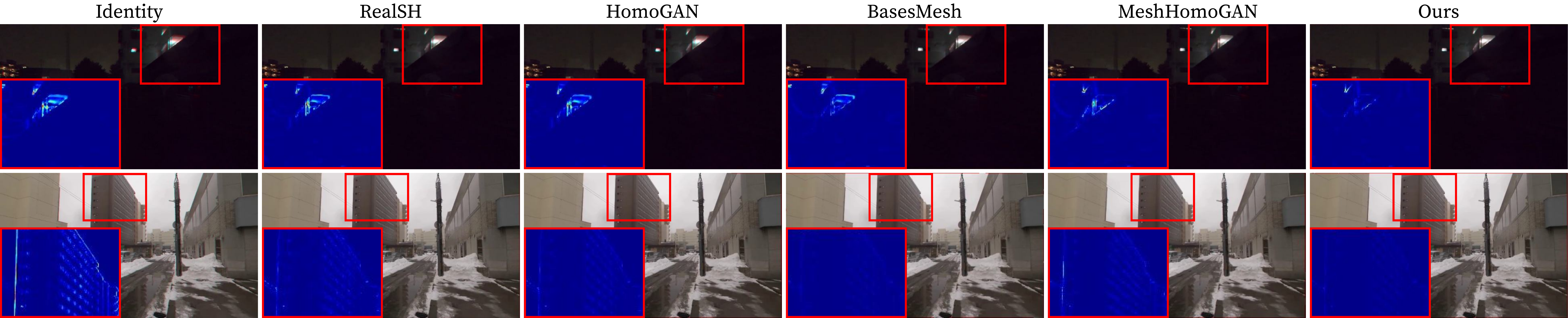}
\end{center}
  \caption{Qualitative results of CamFlow and methods from each category (i.e., supervised, unsupervised, and multi-homography) on the CAHomo test set~\cite{CA-Unsupervised2020}. The images are generated by superimposing the warped source images on the target image. Error-prone regions are highlighted with red boxes, which are further converted into alignment heatmaps for better distinction when zoomed in.}
\label{fig:cahomo_qualitative}
\vspace{-1em}
\end{figure*}

\begin{figure*}[th]
\begin{center}
  \includegraphics[width=1\linewidth]{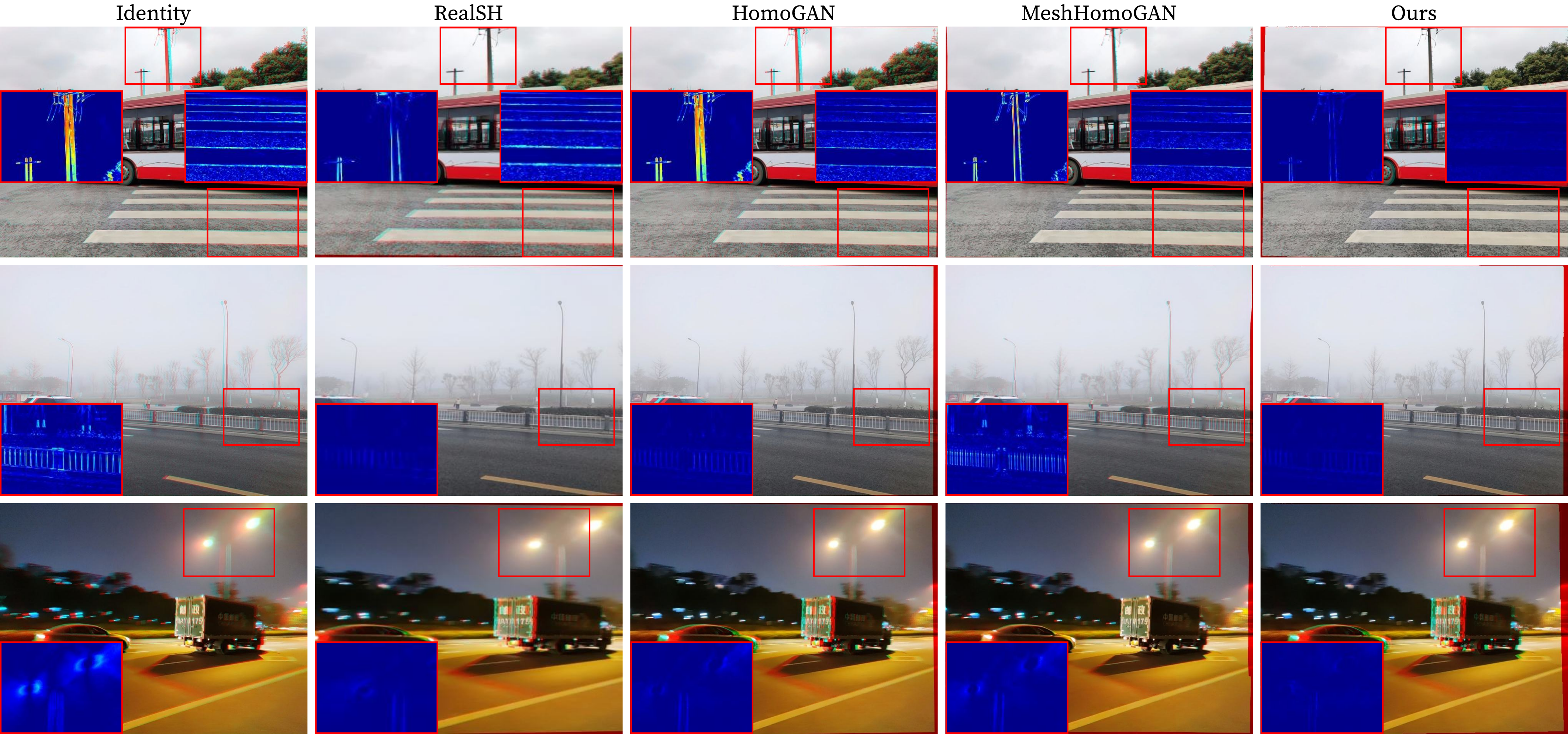}
\end{center}
  \caption{Qualitative results of our method and the best-performing generalizable methods from learning-based categories (i.e., supervised, unsupervised, and multi-homography) on the GHOF test set~\cite{gyroflow+}. The examples are arranged from top to bottom and include RE, FOG, and LL. The images are generated by superimposing the warped source images on the target image. Error-prone regions are highlighted with red boxes, which are further converted into alignment heatmaps for better distinction when zoomed in.}
\label{fig:ghof_qualitative}
\end{figure*}

\subsubsection{Sparse camera motion.}

\noindent\textbf{CAHomo Benchmark.} We use point matching error (PME) to evaluate sparse homography alignment. As shown in Table~\ref{tab:Compare_CAunsup}, CamFlow achieves the best average error among unsupervised methods and is competitive with recent supervised baselines. It also remains stable in low-texture, low-light, and foreground-rich scenes, where feature matching or single-homography models are more easily affected.

\noindent\textbf{GHOF Benchmark.} Table~\ref{tab:Compare_GHOF} reports sparse background PME on GHOF. Traditional feature or geometry methods remain strong when reliable matches exist, while learning-based methods provide better overall robustness across adverse conditions. In zero-shot testing, CamFlow obtains the best average PME among learning-based methods (1.23), outperforming RealSH (1.72), DMHomo (1.75), and HomoGAN (1.95). The gain is most visible in difficult subsets, e.g., CamFlow reduces the low-light error from 4.16/3.95 to 2.69 compared with DMHomo/HomoGAN, and reduces the snowy-scene error from 2.74/3.02 to 0.93.

3D foundation and optical-flow models are included as additional references, but Table~\ref{tab:Compare_GHOF} shows that dedicated 2D camera-motion modeling remains more reliable on GHOF.
CamFlow+ requires camera intrinsics and depth for depth-translational bases. Since CAHomo lacks intrinsics, we train CamFlow+ on GHOF, which provides intrinsics, using VGGT-estimated depth. For a fair adaptation comparison, we also refine the strongest trainable baselines on GHOF. Fine-tuning improves all selected methods: HomoGAN, DMHomo, and CamFlow decrease to 0.79, 0.70, and 0.64 PME, respectively, while CamFlow+ achieves the best average PME of 0.51.

\begin{figure}[t]
    \centering
    \noindent \includegraphics[width=\linewidth]{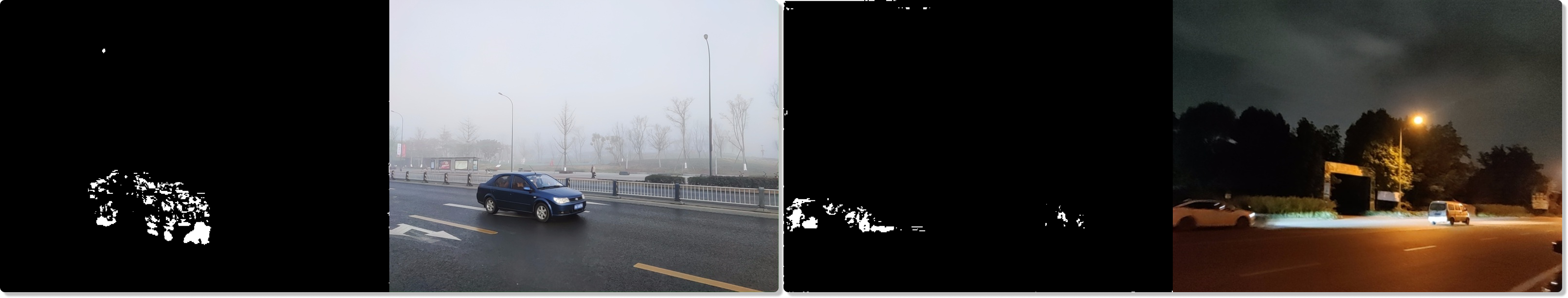}
    \caption{Visualization of foreground masks and corresponding original images. The predicted masks successfully highlight dynamic objects.}
    \label{fig:mask}
\end{figure}

\begin{figure}[t]
    \centering
    \includegraphics[width=1\linewidth]{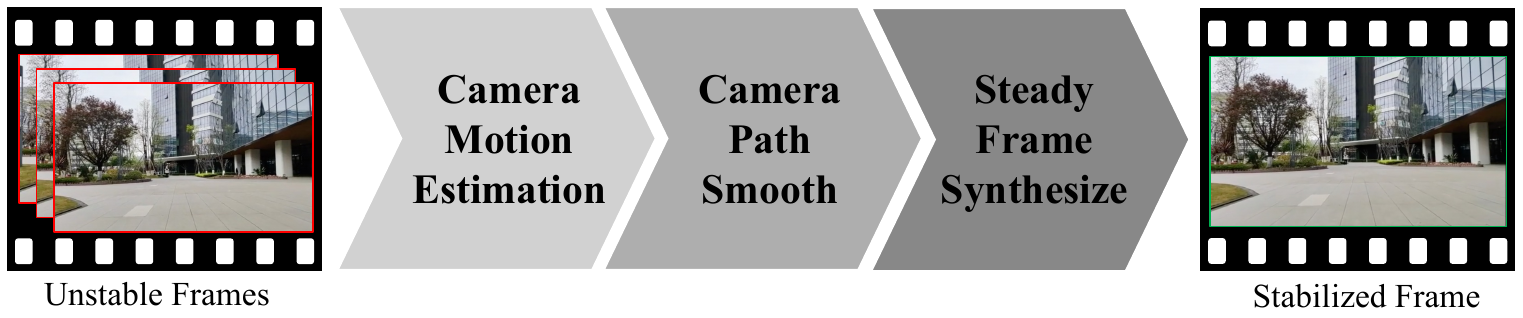}
    \caption{{A typical DVS pipeline.} Three stages: \emph{motion estimation}, \emph{motion smoothing}, and \emph{video rendering}. We keep the last two fixed and vary only the first.}
    \label{fig:dvs_ppl}
\end{figure}

\begin{figure*}[th]
\begin{center}
  \includegraphics[width=1\linewidth]{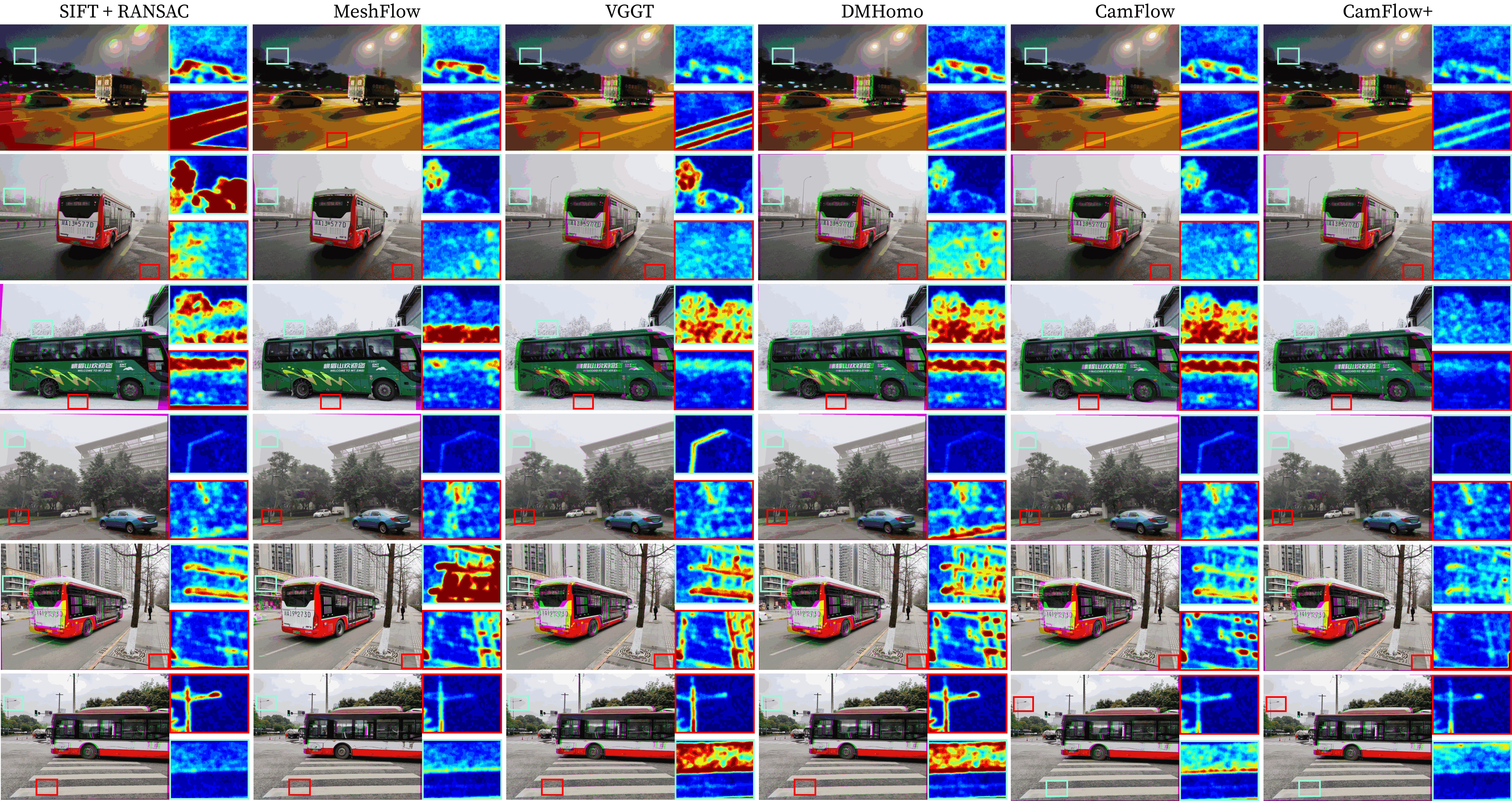}
\end{center}
  \caption{{More visual comparisons on GHOF. We use color-ghosting overlays and alignment heatmaps to visualize registration quality, where weaker ghosting and lower heatmap responses indicate better alignment. Green boxes highlight background regions, while red boxes indicate foreground regions with stronger parallax sensitivity.}}
\label{fig:ghof_ft_qualitative}
\end{figure*}

\subsubsection{Dense camera motion.}

{Table~\ref{tab:GHOF-Cam-Quan} reports dense EPE on GHOF-Cam. Among training-free methods, MeshFlow achieves the best average EPE (2.15), and SIFT remains strong in regular scenes (0.60). For CAHomo-pretrained learning methods, CamFlow obtains the best average EPE (1.10). It is close to HomoGAN on average (1.16), but improves clearly in regular scenes with parallax, reducing EPE from 1.61 to 1.08.}

{3D foundation models still show limited dense camera-motion accuracy on GHOF-Cam, likely due to domain shift in outdoor mobile videos. RAFT and Sea-RAFT achieve strong EPE on valid static pixels (both 0.89 average), confirming the value of dense correspondence pre-training. However, their full predictions also include object motion and occlusion artifacts, as shown in Fig.~\ref{fig:of_remap}, making them hard to use directly for stabilization-like applications. After GHOF fine-tuning, CamFlow+ achieves the best average EPE of 0.50, improving over refined CamFlow (0.64), DMHomo (0.70), and HomoGAN (0.73).}

\subsubsection{Perceptual Quality Metrics.}
{Table~\ref{tab:ghof-cam} reports PSNR, SSIM, and LPIPS after image transformation within camera motion-related regions. These metrics measure whether estimated camera motion produces visually aligned frames. MeshFlow is the strongest training-free baseline, while refined learning methods perform best overall. CamFlow+ achieves the best average perceptual quality, reaching 32.30 PSNR, 0.9320 SSIM, and 0.0487 LPIPS, and improves over CamFlow on all three metrics.}

\subsection{Qualitative Comparison}
\label{subsec:qualit_compare}

Fig.~\ref{fig:cahomo_qualitative} and Fig.~\ref{fig:ghof_qualitative} show qualitative comparisons on CAHomo and GHOF. Following~\cite{LBHomo}, we use color ghosting and alignment heat maps to show registration quality: stronger ghosting or brighter heat-map responses indicate larger misalignment. CamFlow produces fewer artifacts in low-light, parallax, foggy, and dynamic-object scenes, showing stronger robustness than single-homography and mesh-based baselines.

Fig.~\ref{fig:ghof_ft_qualitative} further compares representative methods on GHOF, including traditional baselines (SIFT and MeshFlow), a 3D foundation model (VGGT), and GHOF-refined learning methods (DMHomo, CamFlow and CamFlow+). Blue boxes mark background regions and red boxes mark parallax-sensitive regions. Across adverse weather and regular handheld scenes, CamFlow+ gives more coherent global and local alignment, while competing methods often fail due to unreliable features, local smoothing artifacts, 3D domain shift, or single-plane motion constraints.

\noindent\textbf{Foreground Masks.} Fig.~\ref{fig:mask} shows that the predicted masks successfully highlight dynamic objects, reducing their influence during training.

\begin{figure*}[t]
    \centering
    \includegraphics[width=1\linewidth]{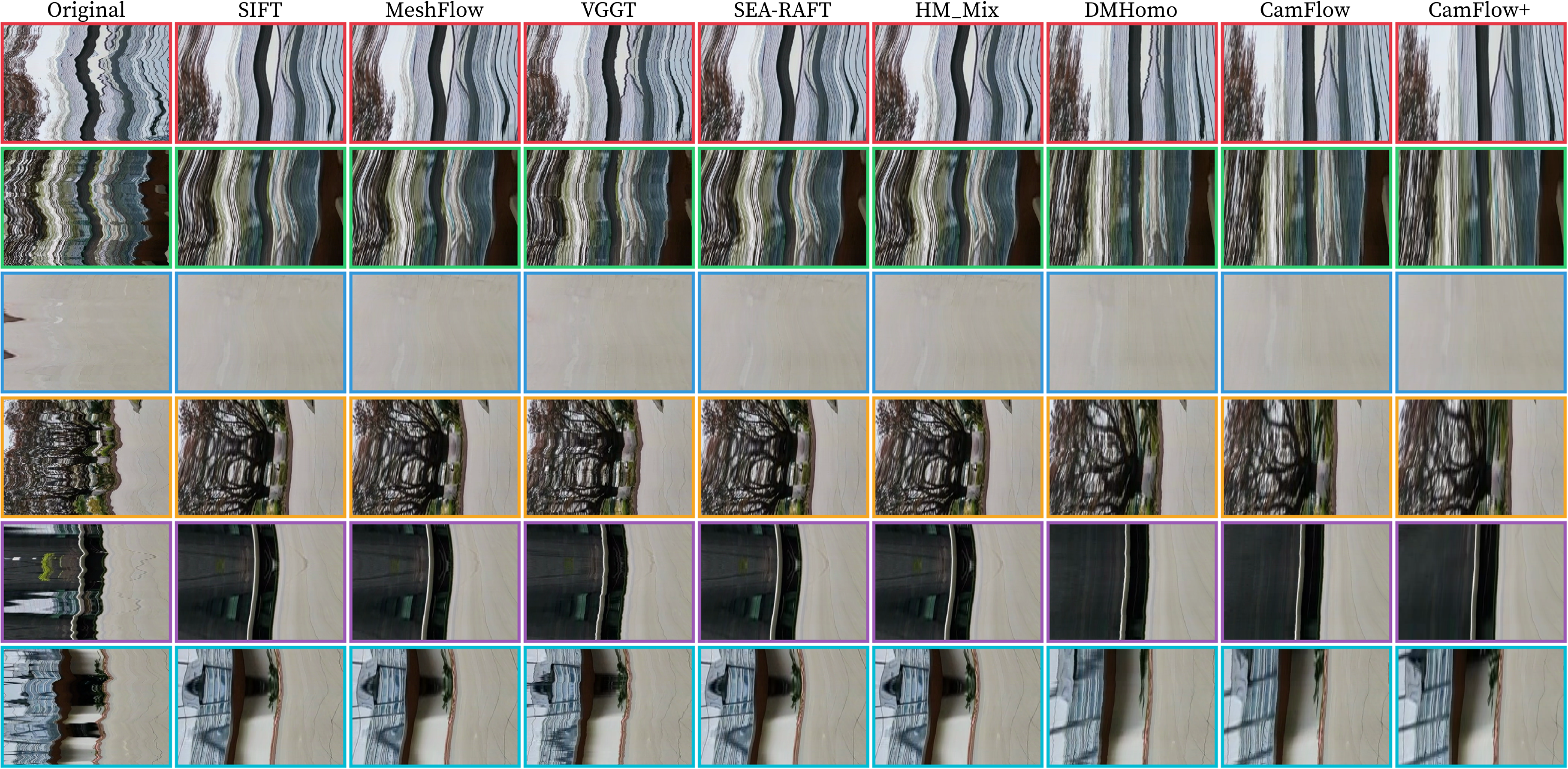}
    \caption{{Spatio-temporal scanline comparison on a hand-held clip.} Uniformly sampled scanlines are stacked along time to form a slice per method. Smooth waveforms indicate low-frequency camera motion; serrated teeth indicate high-frequency jitter. A good stabilizer flattens the teeth while keeping the baseline smooth.}
    \label{fig:dvs-quali}
\end{figure*}

\begin{figure}[h]
    \centering
    \includegraphics[width=1\linewidth]{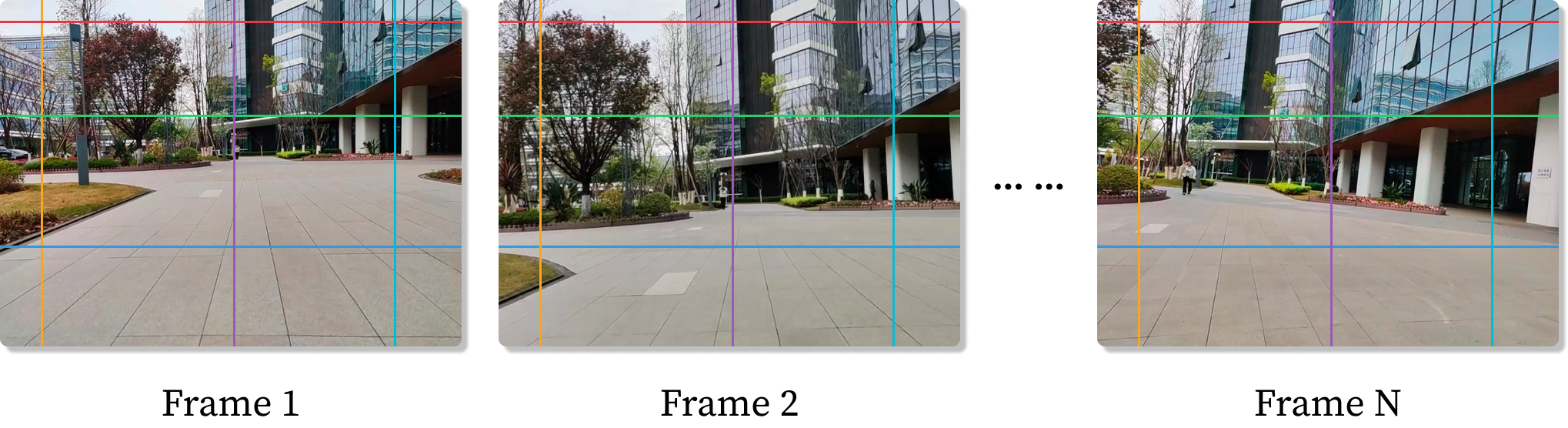}
    \caption{{Scanline selection.} Colored scanlines are placed on scene planes at different depths and span $x$/$y$ camera translations, providing a compact probe of the stabilized motion.}
    \label{fig:dvs-quali-scanline}
\end{figure}

\subsection{Video Stabilization}
\label{sec:application}

\begin{figure*}
    \centering
    \includegraphics[width=1\linewidth]{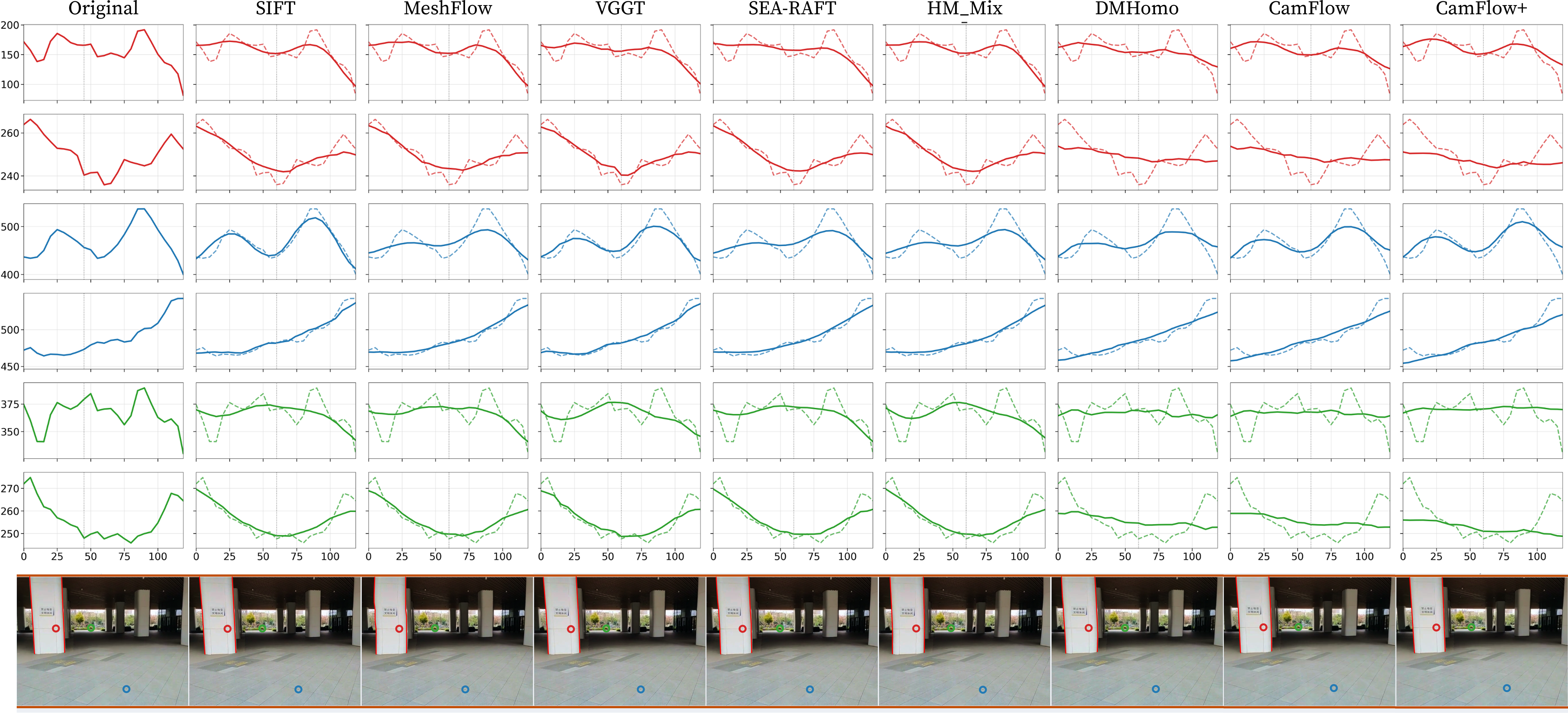}
    \caption{\textbf{Tracked-point trajectories on a pillar scene.} {Red}/{green}/blue points are tracked by KLT with light manual correction. For each method, the top/bottom row shows $x$/$y$ motion; dashed = original, solid = stabilized. The last row traces pillar contours to highlight shape preservation.}
    \label{fig:dvs-quali-video1}
\end{figure*}

{
Digital video stabilization (DVS) is a fundamental computational photography task in mobile imaging. Given a hand-held video, the goal is to suppress high-frequency camera shake while preserving intentional camera motion and avoiding visible warping artifacts. As shown in Fig.~\ref{fig:dvs_ppl}, a typical DVS pipeline comprises three stages: (i) \emph{motion estimation}, which recovers inter-frame camera motion; (ii) \emph{motion smoothing}, which converts the unstable camera trajectory into a visually smooth one; and (iii) \emph{video rendering}, which warps each frame according to the smoothed trajectory to produce the stabilized output. Classical pipelines~\cite{liu2013bundled,SteadyFlow2014} have shown that the accuracy and reliability of the estimated camera motion is a key factor affecting final stabilization quality.
}

{
This setting is thus a natural testbed for CamFlow+.
Here we move beyond benchmark numbers and directly examine whether better camera-motion estimation translates into tangible gains on a practical mobile video-processing task. To isolate the effect of motion estimation, we implement a SteadyFlow-style DVS backend~\cite{SteadyFlow2014} and keep the smoothing and rendering stages fixed, replacing only the first stage with different motion estimators. Concretely, all methods are converted to the same dense optical-flow interface before stabilization: homography-, mesh-, and geometry-based outputs are sampled as per-pixel displacement fields, while dense-flow and CamFlow-based methods already provide this representation. The smoothing parameters, rendering settings, crop window, and zoom strategy are kept identical for all methods. This controlled setup makes the comparison across motion representations directly attributable to the estimator itself.
}

{
We record several new hand-held videos using the same Xiaomi CC9 Pro smartphone as the GHOF benchmark~\cite{gyroflow+} to control camera intrinsics and reduce device-dependent imaging confounders. All clips are captured in regular daily scenes, matching the most application-relevant GHOF category for mobile video stabilization. For comparison, we select representative methods from three groups. \emph{Traditional methods}: the three strongest regular-scene baselines from our benchmark evaluation, namely SIFT~\cite{sift}, MeshFlow~\cite{LiuTYSZ16}, and Homography Mixture (HM-Mix)~\cite{grundmann2012calibration}. \emph{Generic dense-motion or foundation models}: VGGT~\cite{wang2025vggt} and Sea-RAFT~\cite{wang2024sea}. \emph{Learning-based camera-motion methods}: refined DMHomo~\cite{li2024dmhomo}, refined CamFlow, and our refined CamFlow+.
}

{
Because video stabilization is ultimately judged by temporal visual perception, we provide dynamic results in the supplementary material, where stabilization-specific artifacts, such as low-frequency drift, high-frequency jitter, and geometric distortion, can be inspected directly. Since such artifacts and the associated cropping trade-offs are difficult to fully summarize with a single scalar measure, we use these dynamic videos together with a blind preference study as the main application evaluation, and within the manuscript we present two complementary forms of static evidence on two stabilized samples.
}

{
\noindent\textbf{Spatio-temporal scanline visualization.} As shown in Fig.~\ref{fig:dvs-quali-scanline}, we uniformly select a set of scanlines on a representative frame so that they cover scene planes at different depths and span camera translations along both the $x$- and $y$-axes. Collecting these scanlines along the temporal axis yields the spatio-temporal slices compared in Fig.~\ref{fig:dvs-quali}, which compactly summarize camera motion in the stabilized video. In each slice, low-frequency motion manifests as smooth waveforms, while high-frequency jitter manifests as serrated, saw-tooth perturbations. As shown in the \emph{Original} column, the unstabilized video produces clearly wavy low-frequency oscillations \emph{and} dense high-frequency teeth on background-tracking scanlines (e.g., the red one). An ideal stabilizer should preserve a smooth low-frequency baseline while eliminating the high-frequency teeth.
}

{
Reading the slices in Fig.~\ref{fig:dvs-quali} qualitatively, traditional baselines (SIFT~\cite{sift}, MeshFlow~\cite{LiuTYSZ16}, HM-Mix~\cite{grundmann2012calibration}) effectively suppress high-frequency teeth but leave residual low-frequency wobble in the baseline. VGGT~\cite{wang2025vggt} handles high-frequency components poorly, retaining visible jaggedness, whereas Sea-RAFT~\cite{wang2024sea} delivers smoother motion. Overall, the learning-based camera-motion methods produce noticeably flatter low-frequency baselines on this sequence: the scanlines through building facades and the ground plane remain nearly straight, which translates into a perceptually well-locked video. Among them, DMHomo~\cite{li2024dmhomo} still exhibits localized high-frequency jitter; CamFlow attenuates this jitter; and our CamFlow+ achieves the most desirable behavior, simultaneously yielding a stable low-frequency trajectory and effective high-frequency suppression.
}

{
\noindent\textbf{Tracked-point trajectories.} As a complementary view, Fig.~\ref{fig:dvs-quali-video1} presents a second sequence in which the operator stands in place and captures several pillars at different distances while rotating and translating the phone. We use KLT with manual correction to track a handful of representative points, shown in \textcolor{red}{red}, \textcolor{ForestGreen}{green}, and \textcolor{blue}{blue}, and plot, for each method, their $x$-direction (top row) and $y$-direction (bottom row) trajectories. Within each stabilization method, the dashed curve denotes the original (unstabilized) motion and the solid curve denotes the post-stabilization motion.
}

{
Reading the curves qualitatively, the \textcolor{ForestGreen}{green} points on the distant background should ideally be \emph{locked}: CamFlow+ produces solid curves that traverse the dashed reference smoothly with negligible residual jitter, indicating effective stillness of the static scene. The \textcolor{blue}{blue} points on the near foreground are heavily coupled with scene depth and camera translation; overstabilizing them forces local wobble and works against the global stabilization budget. CamFlow+ keeps these points \emph{compliant}, i.e., closer to the original motion, which is in fact the desirable behavior. The \textcolor{red}{red} points on the pillars are the most delicate: pillar boundaries lie on strong depth discontinuities, so aggressive stabilization easily induces wobble and distortion. As highlighted in the last row, where we trace the pillar contours in red for the reader (the effect is more conspicuous in the video), locally adaptive baselines such as MeshFlow~\cite{LiuTYSZ16}, Sea-RAFT~\cite{wang2024sea}, and HM-Mix~\cite{grundmann2012calibration} overshoot and bend, inflate, or otherwise deform the pillars, whereas the single-homography SIFT~\cite{sift} and the 3D-to-2D projective VGGT~\cite{wang2025vggt} preserve shape better at the cost of weaker stabilization. Among the refined learning-based methods, DMHomo~\cite{li2024dmhomo} and CamFlow still induce minor pillar distortions, while CamFlow+ strikes the most favorable balance between shape preservation and motion stabilization. We attribute this improvement to the explicit depth induction and the techniques designed around it, which equip the network with an awareness of scene depth and thereby moderate stabilization aggressiveness near strong depth discontinuities.
}

\begin{figure}[t]
    \centering
    \subfloat[Aggregated top-1 votes.]{
        \label{fig:top1_prefer}
        \includegraphics[width=1\linewidth]{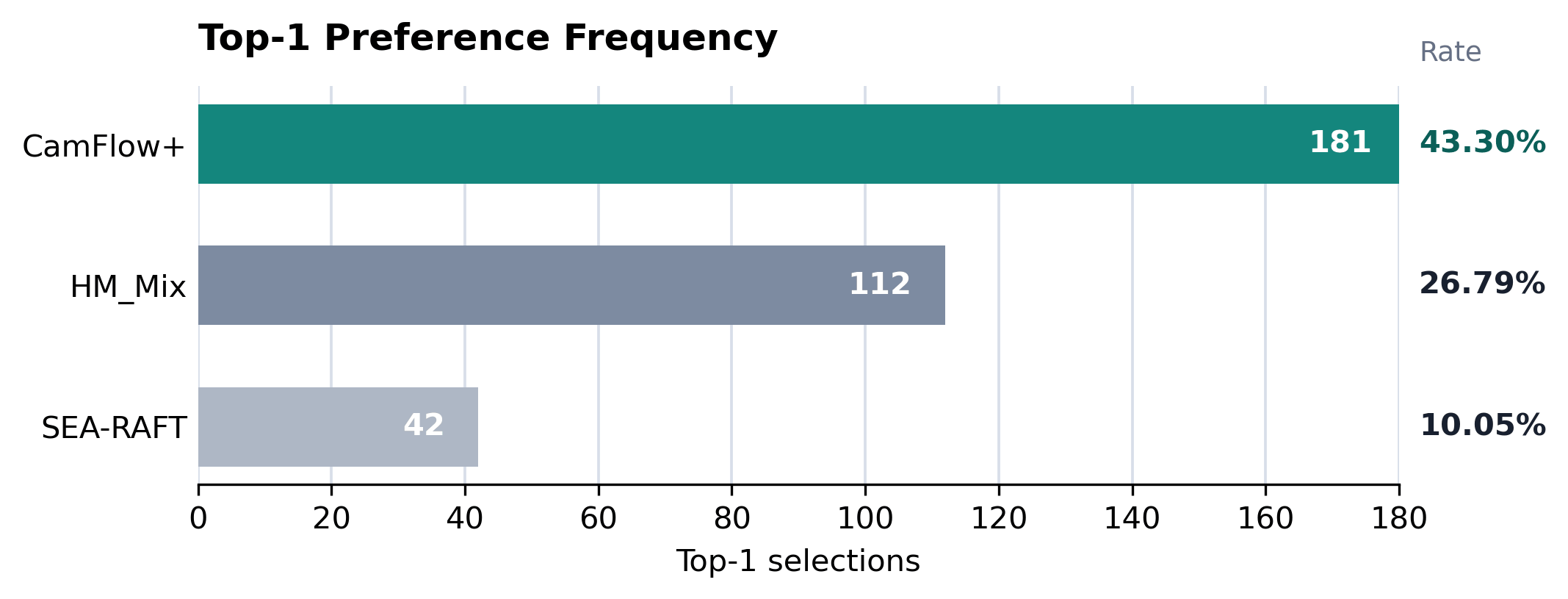}
    }
    \\
    \subfloat[Per-video top-1 votes.]{
        \label{fig:per_video_top1}
        \includegraphics[width=1\linewidth]{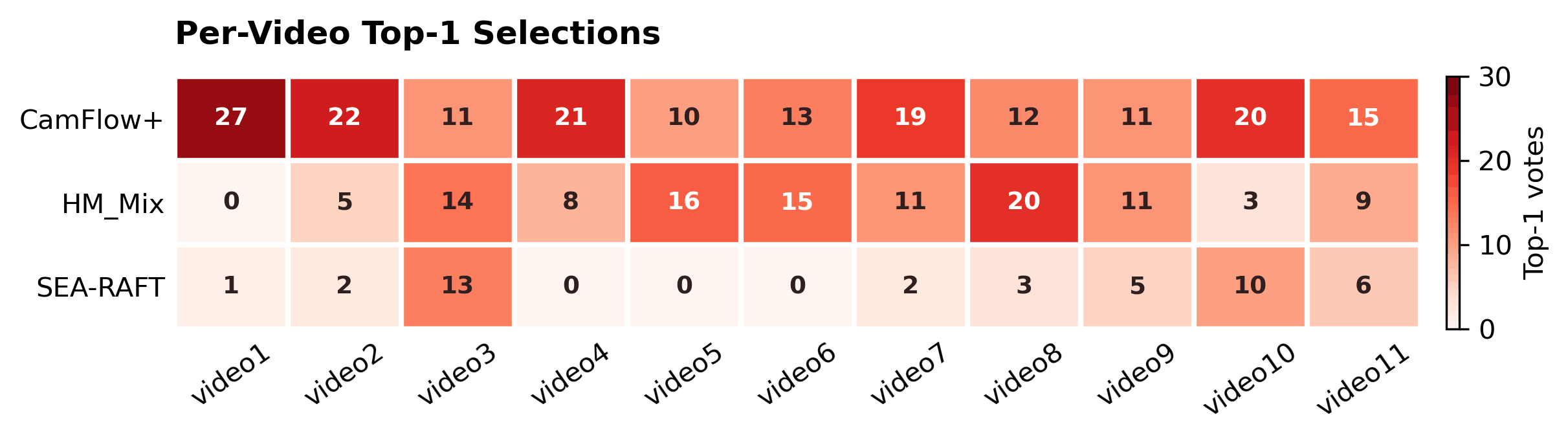}
    }

    \caption{\textbf{Blind user study results.} (a) Total top-$1$ votes received by the three most preferred methods, aggregated over all participants and videos. (b) Per-video top-$1$ votes for the same methods. CamFlow+ is selected as the best stabilized clip in nearly half of all trials and leads on the majority of videos; complete results for all evaluated methods are provided in the supplementary material.}
    \label{fig:user_study}
        \vspace{-2em}
\end{figure}

{
\noindent\textbf{User study.} We further conduct a blind user study to obtain a subjective measure of stabilization. We use $11$ held-out hand-held videos, yielding $38{\times}11=418$ trials from $38$ participants balanced between CV researchers and non-experts. For each video, the eight stabilized results (SIFT, MeshFlow, HM-Mix, VGGT, Sea-RAFT, refined DMHomo, CamFlow, CamFlow+) together with the \emph{Original} clip are arranged into a $3{\times}3$ grid, played synchronously and looped, with grid positions randomly permuted per video and per participant and no method labels shown. Each participant evaluates a grid by an elimination protocol: they progressively remove the clip they find least pleasant or stable until one clip remains. The final remaining clip gives the top-$1$ preference for that trial. Because the full elimination order is recorded, the last three non-eliminated clips also define a top-$3$ inclusion set; we use this only as a coarse high-preference-group statistic, since the protocol does not require a strict ordering among the final candidates.

As shown in Fig.~\ref{fig:top1_prefer}, which reports the three most preferred methods, CamFlow+ receives $181/418$ top-$1$ votes ($43.30\%$), clearly ahead of HM-Mix ($112/418$, $26.79\%$) and Sea-RAFT ($42/418$, $10.05\%$).
The top-$3$ inclusion analysis further shows that both CamFlow+ ($292/418$, $69.86\%$) and HM-Mix ($296/418$, $70.81\%$) are frequently retained as acceptable candidates, while the top-$1$ preferences favor CamFlow+ as the final winner. Sea-RAFT is included in the top-$3$ set in $141/418$ trials ($33.73\%$).

}

{
Together with the supplementary videos, the scanline visualizations, trajectory diagnostics, and the user study support that CamFlow+ translates into tangible improvements in the final DVS output.
}

\subsection{Ablation Studies}
\label{sec:abla_study}

\subsubsection{Motion Basis}
\begin{table}[t]
    \caption{Effect of the number and type of motion bases. All variants are pretrained on CAHomo and tested zero-shot on GHOF/GHOF-Cam. CAHomo and GHOF report PME, while GHOF-Cam reports EPE.}
    \label{tab:ablat1}
    \vspace{-2.5em}
    \begin{center}
    \resizebox{1\linewidth}{!}{
    \begin{tabular}{r|ccc|cc}
    \toprule
    & CAHomo & GHOF & GHOF-Cam & Params & Inference Time \\
    \midrule
    8 Bases & 0.37  & 1.68  & 1.45 &2.658M & 76.42ms \\
    12 Bases & 0.36  & 1.54  & 1.23 &2.658M &75.38ms \\
    \midrule
    24 Bases & 0.33  & 1.23 & 1.10 &2.660M &79.63ms \\
    200 Bases  & 0.33  & 1.27  & 1.07 &2.677M &99.28ms \\
    \bottomrule
    \end{tabular}}
    \end{center}
\end{table}

In Table~\ref{tab:ablat1}, we isolate the effect of basis-set design under a fixed CAHomo pretraining protocol. All variants are trained on CAHomo and then evaluated on the CAHomo test set, GHOF, and GHOF-Cam; the latter two benchmarks are tested in a zero-shot manner without GHOF-domain fine-tuning. The first two rows compare purely physical basis sets, showing that increasing the number of homography-derived physical bases from 8 to 12 consistently improves both sparse PME and dense EPE. The 24-basis model further augments the 12 physical bases with 12 stochastic bases, reducing GHOF PME from 1.54 to 1.23 and GHOF-Cam EPE from 1.23 to 1.10, which indicates better transfer to unseen mobile-camera motion. Expanding the stochastic component more aggressively to 200 total bases (12 physical and 188 stochastic) brings only a small additional gain on GHOF-Cam, slightly degrades GHOF PME, and increases inference time from 79.63 ms to 99.28 ms. We therefore use the 24-basis configuration as the default CamFlow setting for a better accuracy-efficiency trade-off.

\subsubsection{Hybrid Probabilistic Loss}
\begin{table}[t]
    \caption{Ablation study on the probabilistic loss components using the 24-basis CamFlow setting.}
    \label{tab:abla2}
    \vspace{-1em}
    \begin{center}
    \resizebox{0.8\linewidth}{!}{
    \begin{tabular}{cc|ccc}
    \toprule
      $\ell_{NLL_m}$ & $\ell_{NLL_f}$ & CAHomo & GHOF & GHOF-Cam \\
    \midrule

     \Checkmark &  & 0.41  & 2.21  & 2.13  \\
      & \Checkmark & 0.36  & 1.58  & 1.42  \\
     \Checkmark & \Checkmark & 0.33  & 1.23  & 1.10  \\

    \bottomrule
    \end{tabular}}
    \end{center}
    \vspace{-2em}
\end{table}
Table~\ref{tab:abla2} evaluates the hybrid probabilistic loss under the same 24-basis CamFlow setting, which consists of 12 homography-derived physical bases and 12 stochastic bases. Using only motion loss yields limited generalization performance, because pseudo motion labels provide approximate supervision. The feature consistency loss alone performs substantially better, particularly on zero-shot datasets, consistent with findings from prior unsupervised methods. However, our hybrid approach combining both losses achieves the best results across all benchmarks, with improvements on GHOF-Cam. This complementary trend indicates that motion labels provide coarse guidance, while feature consistency enables fine-grained refinement for more accurate and generalizable camera motion estimation.

{
\subsubsection{Depth-Aware Design}
\label{sec:abla_depth_aware}

CamFlow+ adds two depth-aware components: depth-translational bases and depth-aware smoothness.
We ablate them sequentially: first adding the bases, then adding depth-aware smoothness to the full model.
We report dense EPE on GHOF-Cam valid static pixels and sparse PME on the corresponding GHOF annotations.

\noindent\textbf{Depth-translational bases.}
In the first step of the sequential ablation, the depth-aware smoothness loss is \emph{disabled for both variants} in Table~\ref{tab:ablation_depth_basis}, so the two configurations differ only in whether depth-translational bases are used. Under this controlled setting, adding the bases consistently reduces EPE across all six subsets, with the average EPE dropping from 0.64 to 0.49 ($+23.4\%$) and the largest relative gain in regular (RE) scenes ($+43.7\%$, from 0.71 to 0.40). This indicates that, equipped with explicit depth-aware bases, the network is able to better model the depth-induced local parallax that dominates pixel-wise flow accuracy, which is the regime that regular scenes with rich structural depth benefit the most from. PME also improves on five of the six subsets, but the gain is smaller and the regular subset shows a slight regression ($-16.4\%$).

\noindent\textbf{Depth-aware smoothness.}
In the second step, we fix the full CamFlow+ architecture and replace only the smoothness loss in Table~\ref{tab:ablation_depth_smooth}. Conventional smoothness terms used by optical-flow methods rely on \emph{image} edges, which are less suitable for camera-motion estimation because RGB edges mix true depth discontinuities with texture and lighting changes. Our depth-aware smoothness instead uses estimated \emph{depth-map} edges as a geometry-oriented cue for regularizing translation-induced flows, aligning the regularization more closely with depth-dependent camera motion. As a result, the proposed term keeps dense EPE nearly unchanged on average ($0.49$ to $0.50$): most subsets vary within about $4\%$, while FOG is a positive exception with a $15.9\%$ reduction. It also substantially improves sparse camera-motion accuracy measured by PME ($+15.6\%$ on AVG and up to $+51.8\%$ on RE), with improvements on most subsets and an essentially unchanged RAIN result at the reported precision ($0.37$ to $0.37$).

\begin{table}[t]
\caption{
{Ablation study on depth-translational bases. Both variants are fine-tuned on GHOF. The upper block uses the 24-basis CamFlow setting, while the lower block adds 12 depth-translational bases. PME is measured on GHOF, and EPE is measured on GHOF-Cam.}
}
\label{tab:ablation_depth_basis}
\vspace{-1em}
\centering
\resizebox{\linewidth}{!}{
\begin{tabular}{c|c|cccccc}
\toprule
{Depth-Trans. Bases}
& {Metric}
& {AVG} & {RE} & {FOG} & {LL} & {RAIN} & {SNOW} \\
\midrule
\multirow{2}{*}{--}
& {EPE $\downarrow$}
& {0.64} & {0.71} & {0.31} & {1.04} & {0.45} & {0.71} \\
& {PME $\downarrow$}
& {0.64} & {0.55} & {0.39} & {1.15} & {0.41} & {0.68} \\
\midrule
\multirow{2}{*}{\checkmark}
& {EPE $\downarrow$}
& {0.49} & {0.40} & {0.25} & {0.85} & {0.38} & {0.58} \\
& {PME $\downarrow$}
& {0.60} & {0.64} & {0.35} & {1.06} & {0.37} & {0.59} \\
\bottomrule
\end{tabular}
}
\vspace{-2em}
\end{table}

\section{Conclusion}
This paper studies 2D camera-motion estimation from the perspective of motion-basis representation. Starting from the conference version, we analyze why direct flow-field addition of homography-induced motions can still be useful despite the non-linear composition of true projective transformations. This analysis motivates a higher-dimensional hybrid basis space that combines physical homography bases with stochastic bases, improving the ability to represent complex camera-induced motion beyond a fixed 8-dimensional homography basis.
In this journal extension, CamFlow+ further connects the 2D motion representation with the 3D-to-2D camera-motion projection. The projection relation shows that camera translation and scene depth are key factors behind depth-dependent parallax. Based on this observation, we extend the physical basis family with depth-translational bases derived from camera intrinsics and depth, and introduce a depth-aware smoothness term to regularize translation-induced flows.
Extensive experiments on CAHomo, GHOF, and the previously introduced GHOF-Cam benchmark show that the proposed representation improves both sparse and dense camera-motion estimation, especially in scenes with parallax and depth variation. We extend the evaluation to digital video stabilization, where replacing the motion-estimation module leads to better stabilized results in visual diagnostics and a blind user study. These results indicate that explicit 2D camera-motion modeling is useful not only for benchmark alignment, but also for practical mobile video-processing applications.

\begin{table}[t]
\caption{
{Ablation study on depth-aware smoothness. Both variants are fine-tuned on GHOF using CamFlow+ with depth-translational bases. The upper block uses standard smoothness, while the lower block uses the proposed depth-aware smoothness. PME is measured on GHOF, and EPE is measured on GHOF-Cam.}
}
\label{tab:ablation_depth_smooth}
\vspace{-1em}
\centering
\resizebox{\linewidth}{!}{
\begin{tabular}{c|c|cccccc}
\toprule
{Smoothing Mode}
& {Metric}
& {AVG} & {RE} & {FOG} & {LL} & {RAIN} & {SNOW} \\
\midrule
\multirow{2}{*}{standard}
& {EPE $\downarrow$}
& {0.49} & {0.40} & {0.25} & {0.85} & {0.38} & {0.58} \\
& {PME $\downarrow$}
& {0.60} & {0.64} & {0.35} & {1.06} & {0.37} & {0.59} \\
\midrule
\multirow{2}{*}{depth}
& {EPE $\downarrow$}
& {0.50} & {0.42} & {0.21} & {0.86} & {0.40} & {0.59} \\
& {PME $\downarrow$}
& {0.51} & {0.31} & {0.28} & {0.99} & {0.37} & {0.58} \\
\bottomrule
\end{tabular}
}
\vspace{-2em}
\end{table}
}

\section{Limitations and Future Work}

Although CamFlow+ improves 2D camera-motion modeling with hybrid and depth-aware motion bases, the current framework still follows a task-specific training and refinement pipeline. It requires curated training data, pseudo motion supervision, and staged optimization for different evaluation settings. This is a common challenge for learning-based camera-motion estimation, especially when moving from benchmark protocols to open-world mobile videos. With the rapid progress of visual foundation models, an important future direction is to integrate the hybrid-basis representation with more general pre-trained models, reducing task-specific training requirements while keeping the geometric structure that makes the motion representation interpretable and controllable.

Our digital video stabilization study focuses on regular hand-held scenes, where stabilization quality can be assessed under common mobile-capture conditions. The smoothing and rendering stages in the adopted DVS backend are kept fixed and are largely independent of image semantics, which helps isolate the effect of camera-motion estimation. Nevertheless, real-world stabilization also involves broader capture conditions, including adverse weather, low illumination, fast motion, rolling-shutter artifacts, and strong dynamic foregrounds. Extending the application study to more diverse videos and verifying consistent improvements across these conditions is an important direction for future work.

\clearpage

\small
\bibliographystyle{IEEEtran}
\bibliography{main}

\vspace{-5em}

\begin{IEEEbiography}
[{\includegraphics[width=1in,height=1.25in,clip,keepaspectratio]{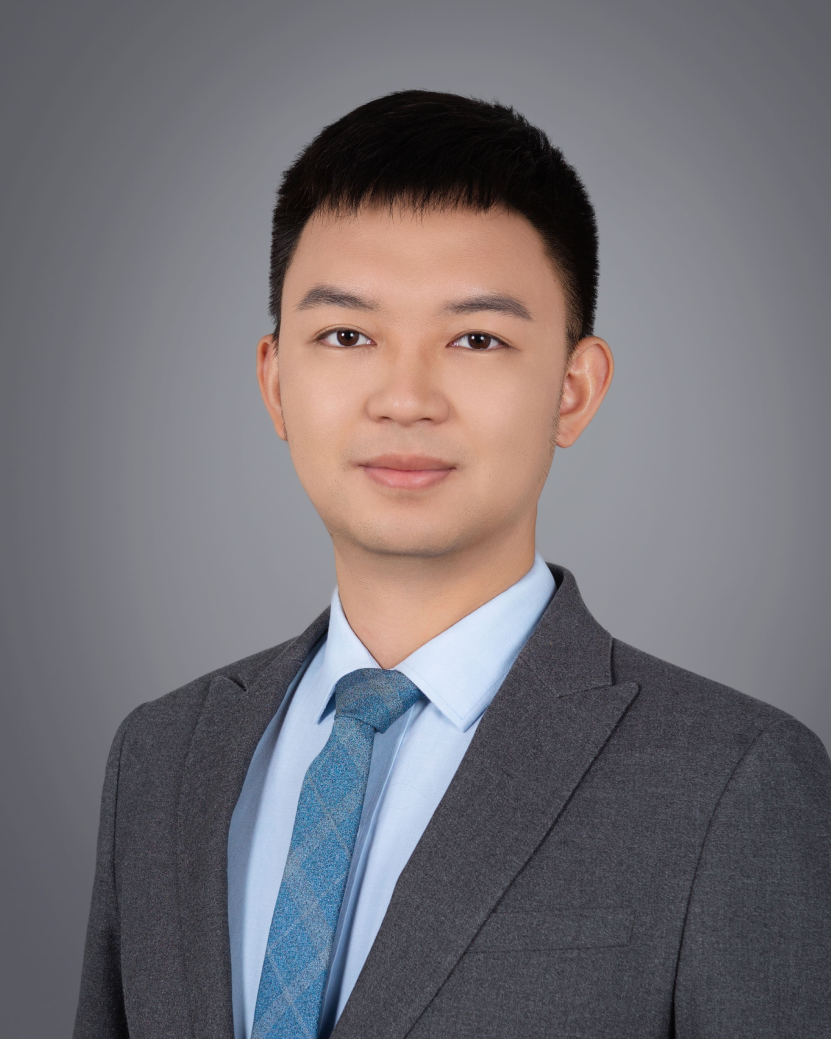}}]{Haipeng Li} received the B.Eng. degree from the University of Electronic Science and Technology of China (UESTC), Chengdu, China, in 2017, and the M.Sc. degree from Institut Mines-Télécom Atlantique Bretagne Pays de la Loire, Brest, France, in 2020. He was a Researcher with Megvii Research from 2020 to 2022. He received the Ph.D. degree from the UESTC, in 2025.
His research interests include computer vision and computer graphics.
\end{IEEEbiography}

\vspace{-5em}

\begin{IEEEbiography}[{\includegraphics[width=1in,height=1.25in,clip,keepaspectratio]{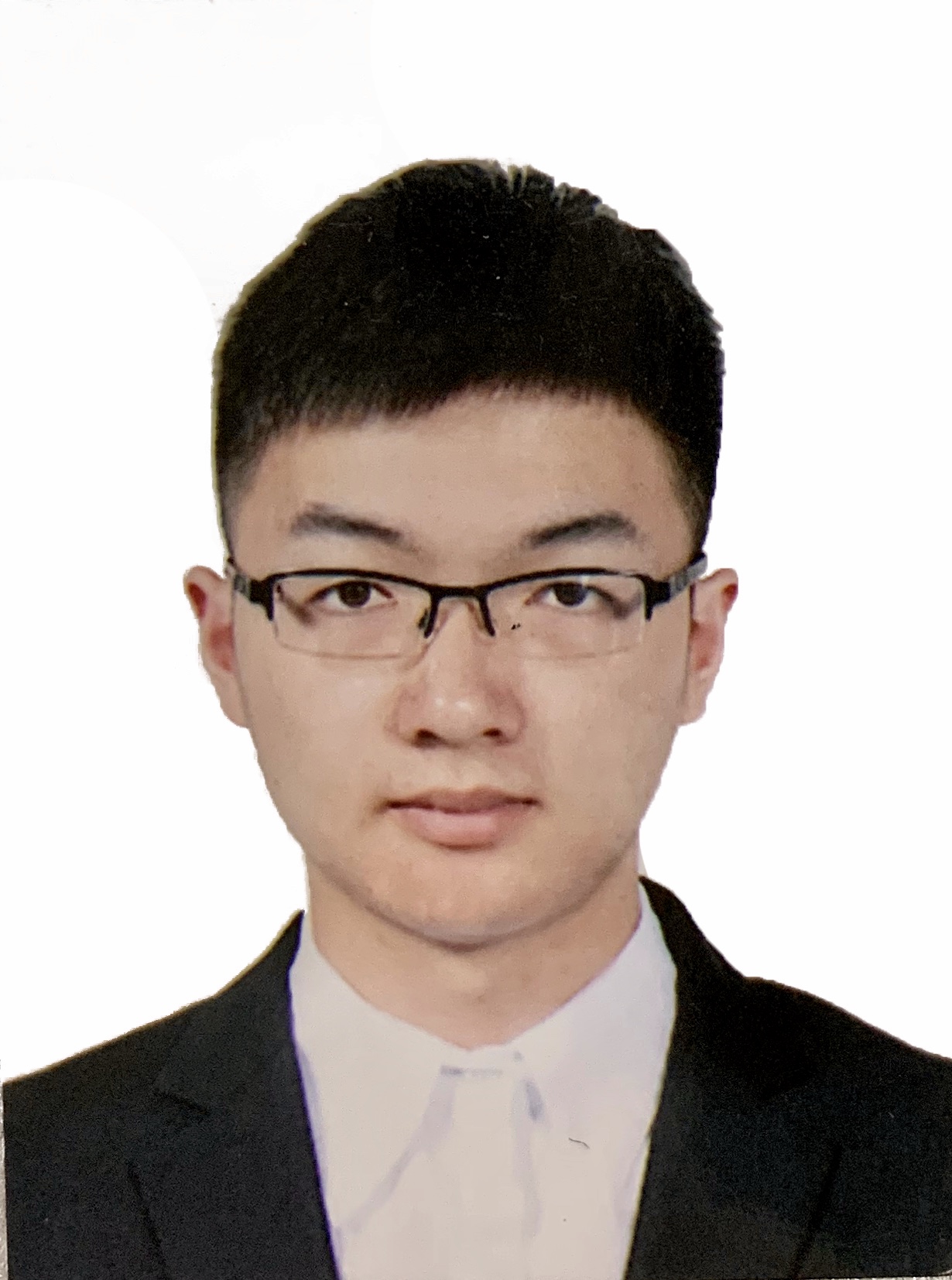}}]{Zhen Liu}
    received the B.S. and M.S. degrees in the College of Computer Science, Sichuan University, Chengdu, China, in 2018 and 2021, respectively. He is currently pursuing the Ph.D. degree with the School of Information and Communication Engineering, UESTC. He was a Researcher with Megvii Research, Chengdu, from 2021 to 2024. His research interests include computer vision and computer graphics. He has served as a reviewer for TPAMI, TIP, IJCV, SIGGRAPH Asia, NeurIPS, ICML, AAAI, CVPR, ICCV, ECCV, etc.
\end{IEEEbiography}

\vspace{-5em}

\begin{IEEEbiography}
[{\includegraphics[width=1in,height=1.25in,clip,keepaspectratio]{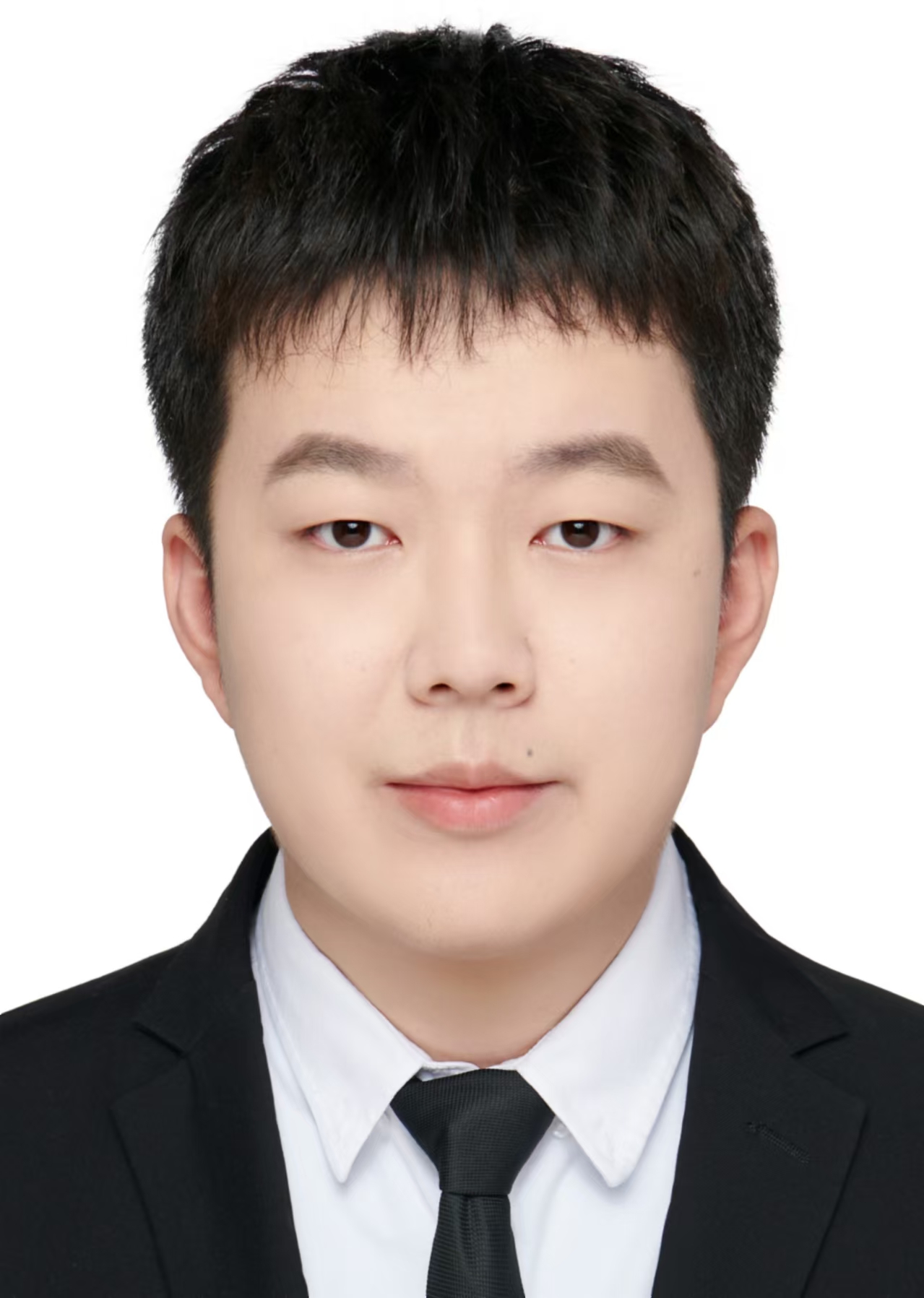}}]{Zhanglei Yang} received the BEng degree from the University of Electronic Science and Technology of China (UESTC), Chengdu, China, in 2022 and the MSc degree from The University of Hong kong, Hong kong, China, in 2024. He has been an intern since 2023 in Megvii Research Chengdu. His research interests include generative models and computer vision.
\end{IEEEbiography}

\begin{IEEEbiography}
[{\includegraphics[width=1in,height=1.25in,clip,keepaspectratio]{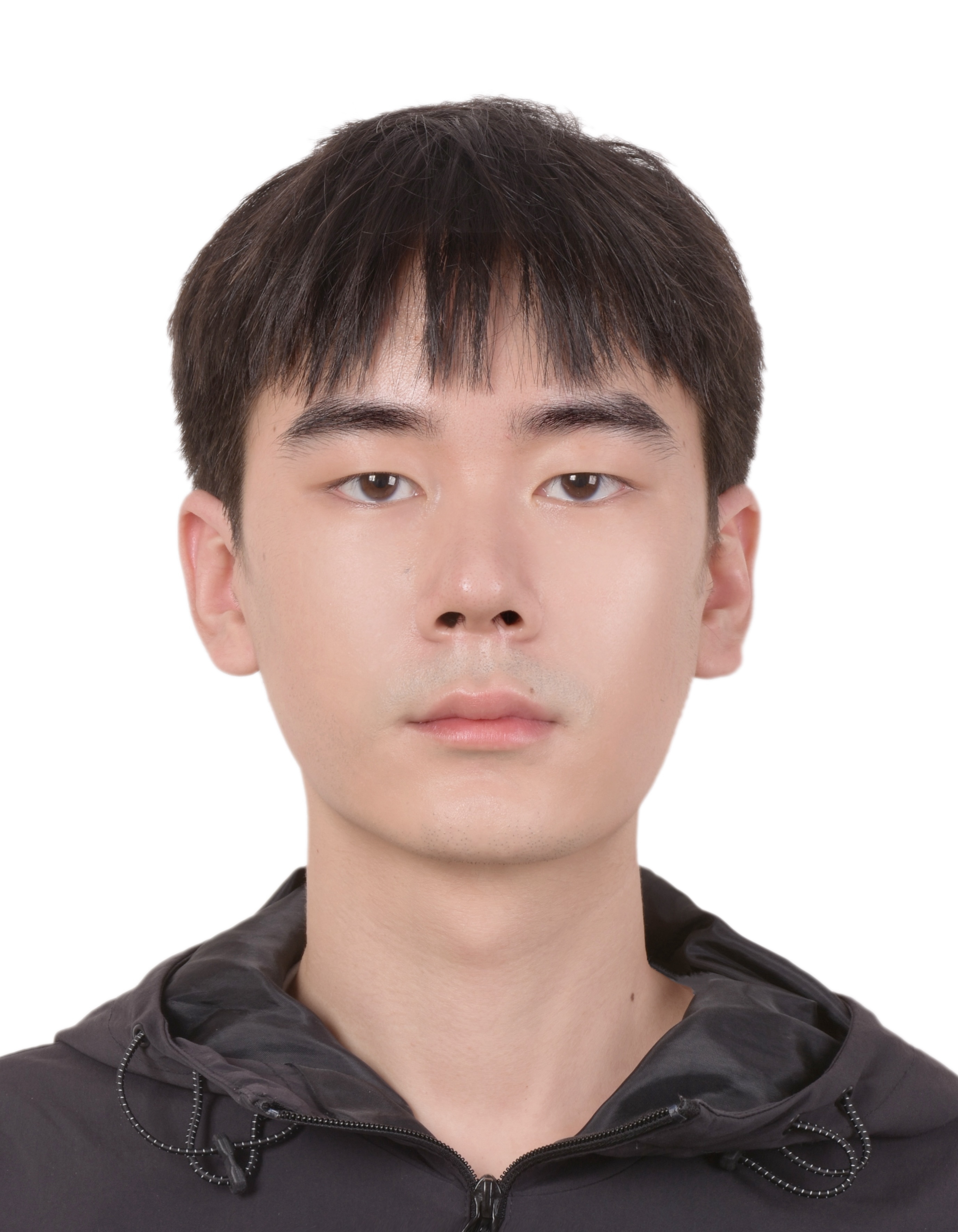}}]{Hai Jiang} received the BEng degree and MSc degree from the School of Aeronautics and Astronautics, Sichuan University, Chengdu, China, in 2016 and 2020, respectively. He is currently working toward the PhD degree with the School of Aeronautics and Astronautics, Sichuan University, and an intern with Megvii Technology. His research interests include computer vision and deep learning.
\end{IEEEbiography}

\vspace{-5em}

\begin{IEEEbiography}
    [{\includegraphics[width=1in,height=1.25in,clip,keepaspectratio]{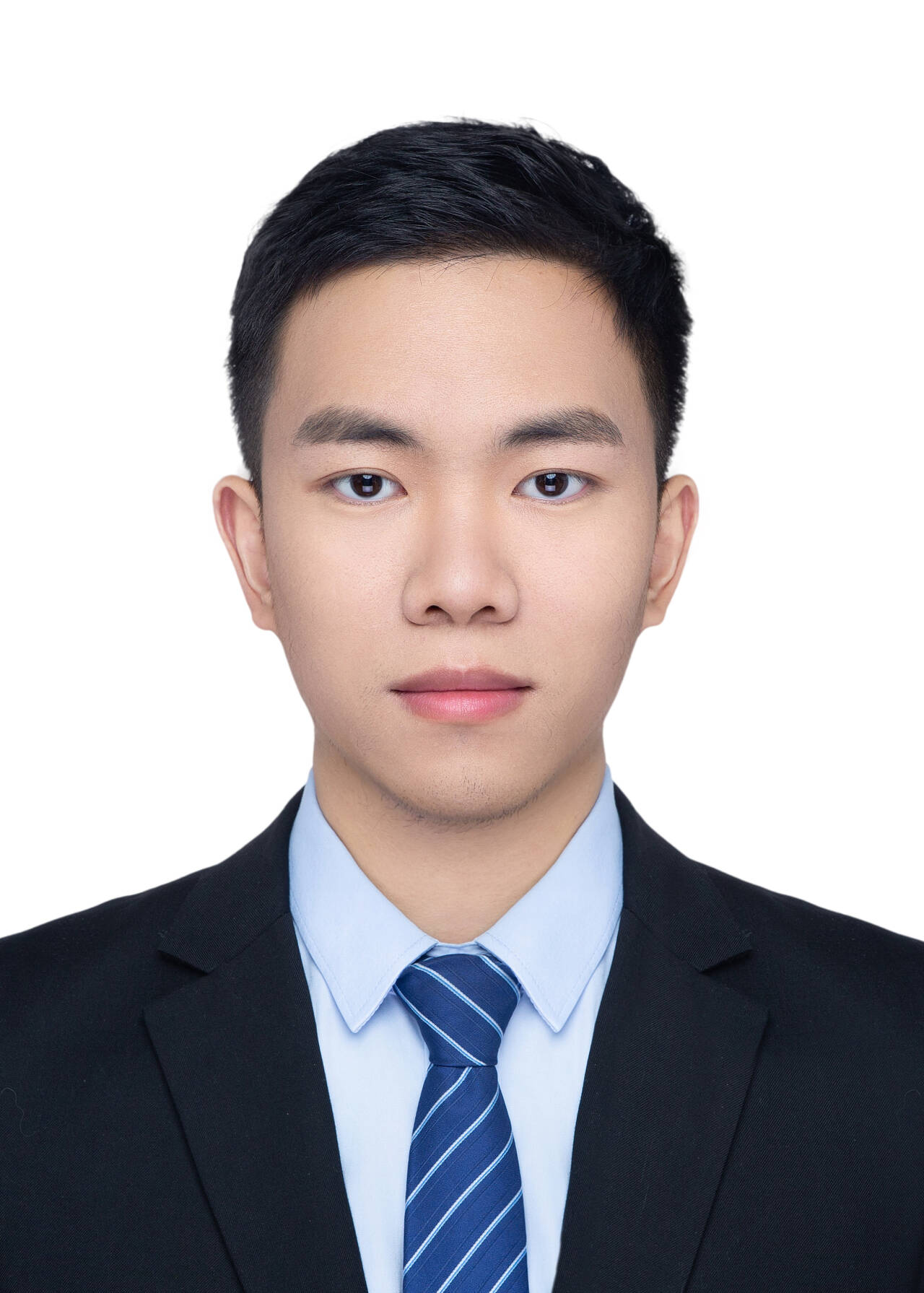}}]{Tianhao Zhou} is an undergraduate student in Yingcai Honors College of the University of Electronic Science and Technology of China. His current research focus on computer vision, deep learning and generative models.
\end{IEEEbiography}

\vspace{-5em}

\begin{IEEEbiography}[{\includegraphics[width=1in,height=1.25in,clip,keepaspectratio]{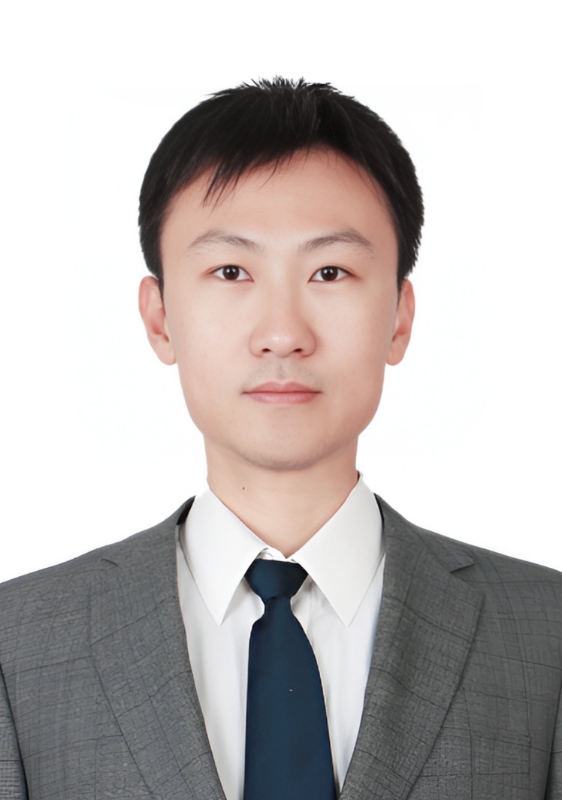}}]{Zhengzhe Liu}
    received the bachelor's degree from Shanghai Jiao Tong University, Shanghai, China, and the M.Phil. and Ph.D. degrees from The Chinese University of Hong Kong, Hong Kong, China. He was a Postdoctoral Research Associate with Carnegie Mellon University, Pittsburgh, PA, USA. He is currently an Assistant Professor with the School of Data Science, Lingnan University, Hong Kong, China. His research interests include computer graphics, embodied AI, generative AI, and computer vision.
\end{IEEEbiography}

\vspace{-5em}

\begin{IEEEbiography}[{\includegraphics[width=1in,height=1.25in,clip,keepaspectratio]{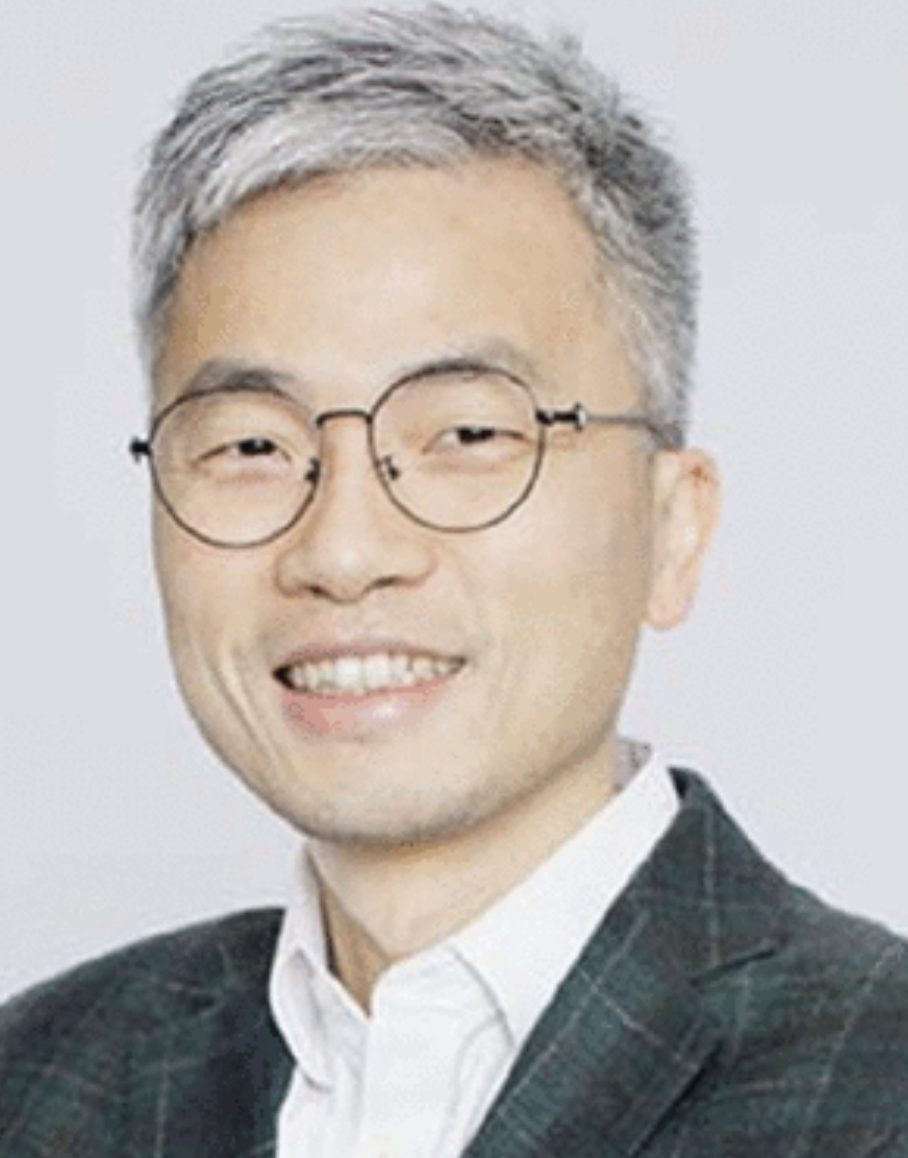}}]{Ping Tan}
 (Senior Member, IEEE) received the bachelor's and master's degrees from Shanghai Jiao Tong University (SJTU), China, in 2000 and 2003, respectively, and the PhD degree from the Hong Kong University of Science and Technology (HKUST), in 2007. He is a professor with the Department of Electronic and Computer Engineering, the Hong Kong University of Science and Technology (HKUST). He was an associate professor with the School of Computing Science, Simon Fraser University (SFU). Before that, he was an associate professor with the National University of Singapore (NUS). His research interests include computer vision, computer graphics, and robotics. He was an editorial board member of the IEEE Transactions on Pattern Analysis and Machine Intelligence, International Journal of Computer Vision, Computer Graphics Forum, and the Machine Vision and Applications, and was the area chairs for CVPR/ICCV, SIGGRAPH, SIGGRAPH Asia, and IROS.
\end{IEEEbiography}

\vspace{-4em}

\begin{IEEEbiography}[{\includegraphics[width=1in,height=1.25in,clip,keepaspectratio]{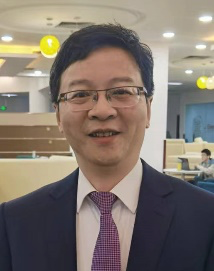}}]{Bing Zeng}
(M'91-SM'13-F'16) received the BEng and MEng degrees in electronic engineering from University of Electronic Science and Technology of China (UESTC), Chengdu, China, in 1983 and 1986, respectively, and the PhD degree in electrical engineering from Tampere University of Technology, Tampere, Finland, in 1991. Currently, he leads the Institute of Image Processing at UESTC and serves as Vice Chair of the Committee for Academic Affairs. He served as an Associate Editor for IEEE TCSVT for 8 years and received the Best Associate Editor Award in 2011. He was elected as an IEEE Fellow in 2016 for contributions to image and video coding.
\end{IEEEbiography}

\vspace{-5em}

\begin{IEEEbiography}[{\includegraphics[width=1in,height=1.25in,clip,keepaspectratio]{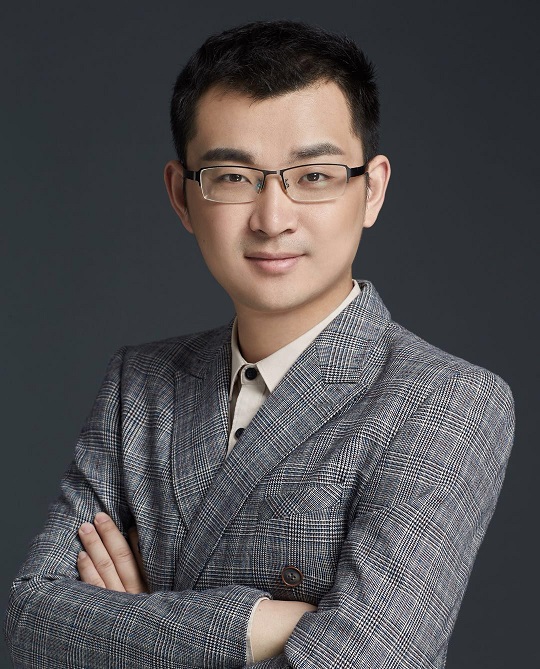}}]{Shuaicheng Liu} (M'14-SM'23) received the Ph.D. and M.Sc. degrees from the National University of Singapore, Singapore, in 2014 and 2010, respectively, and the B.E. degree from Sichuan University, Chengdu, China, in 2008. In 2015, he joined the University of Electronic Science and Technology of China and is currently a Professor with the Institute of Image Processing, School of Information and Communication Engineering, Chengdu, China. His research interests include computer vision and computer graphics.
\end{IEEEbiography}


\end{document}